\documentclass[10pt,twocolumn,letterpaper]{article}

\usepackage[pagenumbers]{cvpr} 
\definecolor{cvprblue}{rgb}{0.21,0.49,0.74}
\usepackage[pagebackref,breaklinks,colorlinks,allcolors=cvprblue]{hyperref}
\usepackage{multirow}
\usepackage{makecell}
\usepackage{color}
\usepackage{relsize}

\newcommand{\zhe}[1]{\textcolor{black}{#1}}

\usepackage{colortbl}

\def\paperID{} 
\def\confName{CVPR}
\def\confYear{2026}

\title{RAID: Retrieval-Augmented Anomaly Detection}

\author{Mingxiu Cai$^{1,}$$^{*}$ \quad Zhe Zhang$^{1,}$$^{*}$ \quad Gaochang Wu$^{1,}$$^{\dagger}$ \quad Tianyou Chai$^{1}$ \quad Xiatian Zhu$^{2}$\\
$^{1}$State Key Laboratory of Synthetical Automation for Process Industries, Northeastern University\\
$^{2}$University of Surrey
}

\begin{document}

\maketitle
\footnotetext{$^*$Equal Contribution.}
\footnotetext{$^\dagger$Gaochang Wu (wugc@mail.neu.edu.cn) is the corresponding author.}
\begin{abstract}
Unsupervised Anomaly Detection (UAD) aims to identify abnormal regions by establishing correspondences between test images and normal templates. Existing methods primarily rely on image reconstruction or template retrieval but face a fundamental challenge: matching between test images and normal templates inevitably introduces noise due to intra-class variations, imperfect correspondences, and limited templates. Observing that Retrieval-Augmented Generation (RAG) leverages retrieved samples directly in the generation process, we reinterpret UAD through this lens and introduce \textbf{RAID}, a retrieval-augmented UAD framework designed for noise-resilient anomaly detection and localization. Unlike standard RAG that enriches context or knowledge, we focus on using retrieved normal samples to guide noise suppression in anomaly map generation. RAID retrieves class-, semantic-, and instance-level representations from a hierarchical vector database, forming a coarse-to-fine pipeline. A matching cost volume correlates the input with retrieved exemplars, followed by a guided Mixture-of-Experts (MoE) network that leverages the retrieved samples to adaptively suppress matching noise and produce fine-grained anomaly maps. RAID achieves state-of-the-art performance across full-shot, few-shot, and multi-dataset settings on MVTec, VisA, MPDD, and BTAD benchmarks. \href{https://github.com/Mingxiu-Cai/RAID}{https://github.com/Mingxiu-Cai/RAID}.
\end{abstract}

\section{Introduction}
\label{sec:intro}

Anomaly detection serves as a cornerstone task in computer vision with applications in industrial quality inspection \cite{01,02}, medical image analysis \cite{03}, and intelligent surveillance \cite{yang2024videoAD}. Given the limited availability and diverse nature of anomaly patterns, recent studies \cite{05,06,guo2025dinomaly} have increasingly gravitated toward Unsupervised Anomaly Detection (UAD) without access to anomalous samples. In parallel, the shift from the traditional one-class-one-model paradigm \cite{patchcore,08} toward unified multi-class UAD \cite{glad,zhao2023omnial,uniad} is steadily enhancing its practical applicability.

\begin{figure}
\begin{center}
\centerline{\includegraphics[width=0.999\linewidth]{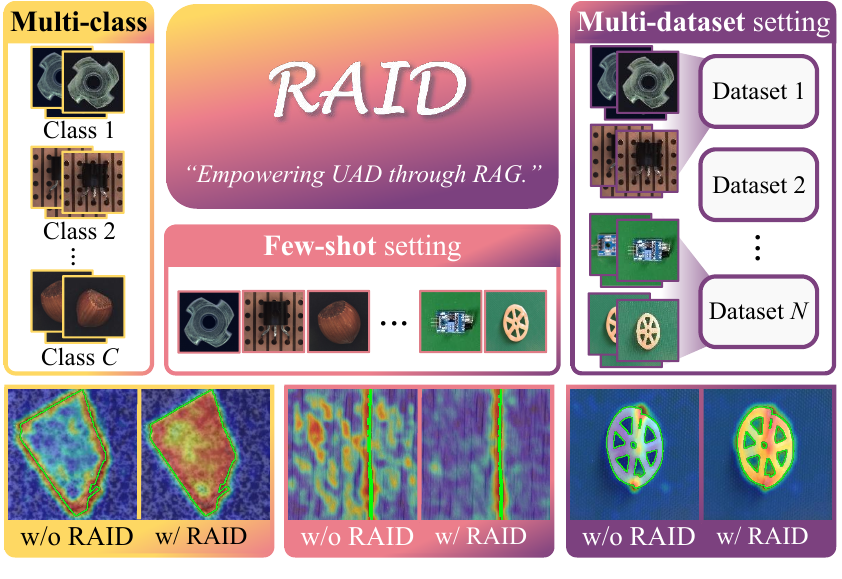}}
\vspace{-2mm}
\caption{We reformulate UAD within the Retrieval-Augmented Generation (RAG) paradigm, effectively reducing retrieval and matching noise and enabling stronger generalization across full-shot, few-shot, and multi-dataset settings.}
\label{teaser}
\end{center}
\vspace{-13mm}
\end{figure}

Existing UAD methods can be broadly categorized into reconstruction- and embedding-based paradigms~\cite{survey1}. Reconstruction-based methods leverage Generative Adversarial Networks (GANs) \cite{17}, Transformers \cite{vitad,guo2025dinomaly,HVQTrans}, or diffusion models \cite{08} to learn the manifold of normal patterns, projecting inputs onto this manifold to obtain normal-looking reconstructions, where discrepancies from the inputs reveal anomalies. \zhe{Embedding-based methods \cite{destseg, uninet, patchguard}, in contrast, bypass explicit reconstruction and instead perform feature-level matching between a query and its corresponding normal templates stored in a memory bank, which may consist of image-level features~\cite{guo2023template}, or patch-level embeddings~\cite{patchcore,Anomalydino,dictas}.} Recently, growing attention has been directed toward leveraging vision foundation models~\cite{clip,dinov2,simeoni2025dinov3} to enhance both paradigms with semantically rich representations, e.g., WinCLIP~\cite{jeong2023winclip}, AnomalyDINO~\cite{Anomalydino}, and Dinomaly~\cite{guo2025dinomaly}. However, these approaches still suffer from unreliable feature matching arising from imperfect reconstructions \cite{survey1} or suboptimal template retrieval, and limited few-shot generalization when adapting to domain-specific categories~\cite{FoundationModel4AD}.

In this paper, we reinterpret UAD through the lens of the Retrieval-Augmented Generation (RAG) paradigm, as illustrated in Fig.~\ref{teaser}. In the era of reasoning-driven artificial intelligence, RAG has emerged as an effective framework to mitigate hallucinations and enhance generalization when models lack sufficient domain-specific knowledge or data~\cite{RAGreviewLLM}. It has been successfully applied to diverse vision tasks~\cite{RAGreview}, including conditional image generation \cite{RADiffusion}, long-tailed image classification \cite{RAC}, and open-vocabulary object detection \cite{RALF}. 
From this perspective, most reconstruction- and embedding-based UAD approaches can be conceptualized as partial realizations of the RAG pipeline, where the model \texttt{retrieves} a normal counterpart (via reconstruction, e.g., GLAD \cite{glad}, memory retrieval, e.g., PatchCore \cite{patchcore}, or teacher-student distillation, e.g., RD++ \cite{tien2023revisiting}) and identifies anomalies via feature matching.
Despite this conceptual alignment, most existing methods overlook the \texttt{generative} reasoning stage of RAG, producing hallucinatory detection noise (e.g., blurred anomaly boundaries and missing subtle anomalies) due to unreliable feature matching.


To address this gap, we propose RAID, a Retrieval-Augmented Industrial anomaly Detection framework that fully integrates the RAG pipeline for UAD, which \texttt{retrieves} normal representations and subsequently \texttt{generates} anomaly maps by jointly reasoning over the retrieved patches and the input patches.
To achieve efficient and scalable retrieval, we design a hierarchical vector database that organizes tokenized templates into three levels of entities, class prototype (category-level concept), semantic prototype (clustered patch token), and instance token (patch token), rather than adopting a flat structure. Compared to existing image-level template retrieval and flat retrieval schemes, this hierarchy facilitates a coarse-to-fine retrieval flow, enabling query tokens to efficiently access semantically relevant template patches with strong contextual consistency for downstream anomaly generation.

We model the generation stage as a guided Mixture-of-Experts (MoE) filtering network designed to mitigate potential matching noise between the input and its multiple retrieved counterparts. Following CostFilter-AD \cite{CostFilterAD}, we first construct a matching cost volume by correlating each input token with its retrieved template tokens. In contrast to previous approaches \cite{Anomalydino,CostFilterAD}, the hierarchical retrieval structure in RAID effectively reduces the matching dimensionality in the cost volume while preserving semantic fidelity. To further mitigate unreliable feature correspondence, the proposed guided MoE filter leverages both the input tokens and the retrieved semantic prototypes as dual guidance, adaptively assigning multiple denoising experts that specialize in distinct semantic and spatial contexts to generate a refined anomaly map that preserves fine-grained anomaly boundaries and subtle anomalies.
Leveraging category- and dataset-agnostic retrieval, RAID injects universal semantic priors into a dynamically activated MoE filter, enabling it to focus on matching-cost denoising and learn category-agnostic anomaly representations. This joint retrieval-generation scheme leads to robust anomaly localization and strong few-shot generalization across unseen categories, as demonstrated in Fig.~\ref{teaser} (bottom).

We summarize our contributions as follows: 
1) We propose RAID, a new paradigm that reconceptualizes UAD within the RAG framework, enabling reliable detection and localization with strong generalization.
2) We introduce a hierarchical vector database that enables a coarse-to-fine retrieval flow, allowing query tokens to efficiently access semantically relevant template tokens with strong contextual consistency.
3) We design the generation stage as a guided MoE filtering network, which dynamically allocates denoising experts to suppress hallucinatory matching noise and improve robustness against diverse anomaly distributions.
4) We extensively evaluate RAID under full-shot, few-shot, and multi-dataset settings, where it consistently outperforms existing methods across multiple benchmarks, demonstrating superior generalization and scalability.


\section{Related works}
\subsection{Unsupervised Anomaly Detection}
Early studies mainly followed the \emph{single-class} paradigm, where models are trained on normal samples to detect anomalies at test time~\cite{draem, patchcore}. With the advent of foundation models and the emergence of new architectures~\cite{vit, sam, dinov2, simeoni2025dinov3}, research has progressed along two directions: (i) \emph{multi-class} detection using a unified model to identify all categories within a dataset~\cite{uniad, HVQTrans, vpdm, glad, CostFilterAD}; and (ii) \emph{few-shot detection} for unknown classes unseen during training~\cite{aprilgan, anomalyclip, adaclip, ucf}, which leverages a small set of auxiliary anomalies without class overlap to the test domain.  
Existing approaches can be broadly grouped into \emph{reconstruction-based} and \emph{embedding-based} families~\cite{survey1, survey2}.
Reconstruction-based pipelines attempt to map anomalous regions back to normal ones and identify discrepancies through similarity or residual analysis. Representative architectures include U-Net~\cite{draem,zhao2023omnial}, Vision Transformer (ViT)~\cite{guo2025dinomaly, dinomalyv2}, Mamba~\cite{mambaad}, Diffusion~\cite{vpdm, diad}, and MoE-based~\cite{moead} frameworks. These models typically integrate pre-trained features with customized decoders~\cite{guo2025dinomaly, mambaad, uniad, HVQTrans, glad} to address the pervasive ``identical shortcut'' issue~\cite{survey1}, which weakens matching similarity or residual contrast.

\begin{figure*}
\begin{center}
\centerline{\includegraphics[width=0.999\textwidth]{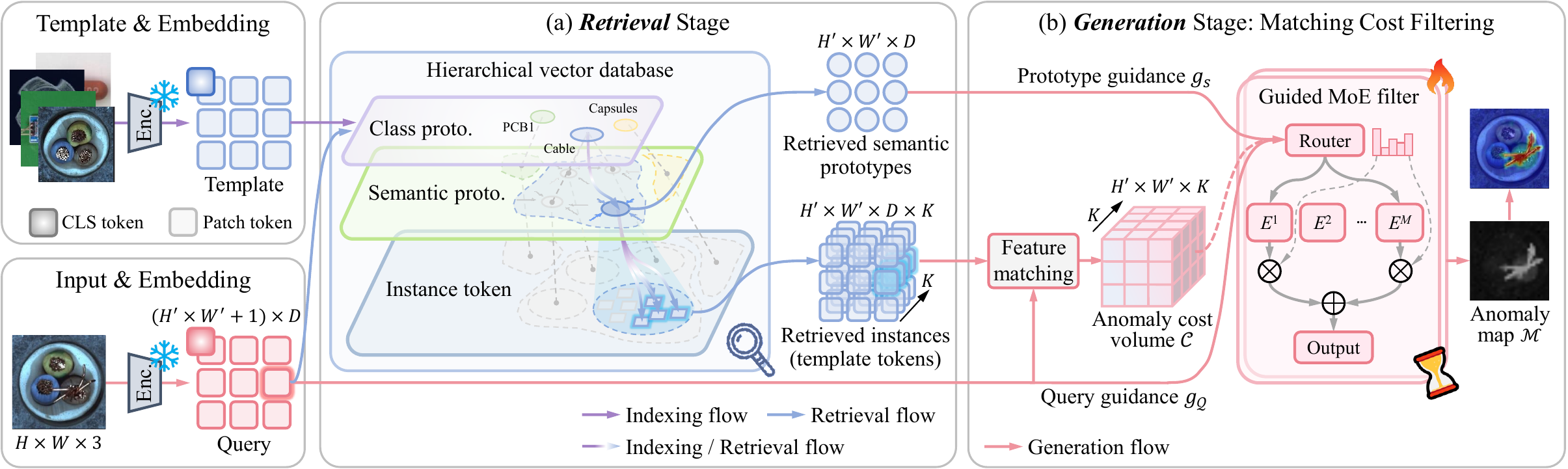}}
\vspace{-2mm}
\caption{Overview of our RAID, which reinterprets UAD within a RAG paradigm. (a) In the \texttt{retrieval} stage, a hierarchical vector database is constructed, indexing tokenized templates into three sequential entity levels: class prototype, semantic prototype, and instance token. This structure allows efficient retrieval flow queried by input tokens. (b) In the \texttt{generation} stage, an anomaly cost volume is built by matching each input token with its retrieved template tokens. A guided MoE filter then dynamically refines this cost volume under the dual guidance of the retrieved semantic prototypes and the input tokens.}
\label{pipeline}
\end{center}
\vspace{-10mm}
\end{figure*}

Embedding-based methods learn either class-specific~\cite{patchcore, simplenet, destseg} or class-agnostic~\cite{dictas, uninet} normal representations, and detect outliers by modeling a normal distribution~\cite{msflow}, maintaining a memory bank~\cite{patchcore} that stores image-level~\cite{guo2023template} or patch-level embeddings~\cite{patchcore,Anomalydino,dictas}. While effective, these methods often suffer from high computational overhead due to large template search spaces~\cite{survey1} or matching noise caused by non-ideal retrieval templates. CostFilter-AD~\cite{CostFilterAD} further introduces a matching-cost filtering module as a plug-in to suppress such noise, yet its performance remains constrained by the initial anomaly cues provided by the host models. Recently, CLIP-based approaches~\cite{jeong2023winclip, anomalyclip, seas, iipad} have been explored for novel-class anomaly detection in the \emph{few-shot} setting, where text embeddings are treated as semantic templates. However, their success heavily depends on the diversity and representativeness of auxiliary real anomaly data.

\subsection{Retrieval-Augmented Generation in Vision}

Retrieval-Augmented Generation (RAG) injects retrievable external knowledge around encoding/decoding to align with model context~\cite{RAGreviewLLM, ref124, ref3}, improving factual consistency, interpretability, and timeliness. Originally developed for language and text understanding~\cite{RAGreviewLLM}, RAG has progressively extended to vision tasks~\cite{RAGreview}. In image understanding, notable directions include retrieval-augmented open-vocabulary object detection~\cite{RALF}, image captioning and generation~\cite{ref120}, medical image segmentation for clinical decision-making~\cite{ref125}, and question answering over visually rich, mixed-format documents~\cite{ref129}.
For visual generation, retrieved image priors are integrated into diffusion and reconstruction pipelines to refine structural and textural fidelity, exemplified by retrieval-augmented text-to-image generation~\cite{ref121}, image restoration~\cite{ref122}, and super-resolution~\cite{ref117}. In video understanding, CadenceRAG~\cite{ref126} demonstrates that retrieval across large video libraries enhances text-grounded video retrieval and question answering, while multi-query and multi-evidence strategies further advance multimodal RAG~\cite{ref9, ref127}.



A conceptually related work is ADSeeker~\cite{adseeker}, which casts anomaly analysis as vision and language question answering that retrieves image-document knowledge to condition an MLLM~\cite{qwen2} for generating textual anomaly descriptions and image-level decisions. In contrast, RAID targets purely visual UAD with pixel-level output (i.e., segmentation) without relying on text Information. It performs hierarchical retrieval over visual exemplars to strike a balance between accuracy and efficiency, and further leverages filtering-based generative reasoning to suppress matching noise and enhance anomaly localization. This design not only clearly distinguishes RAID from existing retrieval-only approaches~\cite{patchcore, guo2023template, Anomalydino, dictas}, but also delivers robust generalization and superior performance across \emph{multi-class}, \emph{few-shot}, and \emph{multi-dataset} UAD settings.

\section{Methodology}

\subsection{Overview}
Given an input (query) image $x_\mathcal{Q}\in \mathbb{R}^{H \times W\times 3}$ (channel, height, and width) or a template (reference) image $x_\mathcal{T}\in \mathbb{R}^{H \times W\times 3}$ prepared for the database, a pre-trained ViT encoder is employed to embed them into patch tokens $\{\mathbf{p}_\mathcal{Q}\}, \{\mathbf{p}_\mathcal{T}\}\in\mathbb{R}^{(H'\times W')\times D}$ and class (CLS) tokens $\mathbf{c}_\mathcal{Q}$, $\mathbf{c}_\mathcal{T}\in\mathbb{R}^{1\times D}$, where $H'$, $W'$, and $D$ denote the resulting spatial resolution and token dimension, respectively. The cosine similarity between a query token vector $\mathbf{p}_\mathcal{Q}$ and a template token $\mathbf{p}_\mathcal{T}$ is defined as: 
\vspace{-2mm}
\begin{equation}\label{eq:cos_sim}
\mathrm{sim}(\mathbf{p}_\mathcal{Q}, \mathbf{p}_\mathcal{T}) = \frac{\mathbf{p}_\mathcal{Q} \mathbf{p}_\mathcal{T}^\top}{\|\mathbf{p}_\mathcal{Q}\| \, \|\mathbf{p}_\mathcal{T}\|},
\vspace{-1mm}
\end{equation}
where $\| \cdot \|$ is the $L_2$ norm. In most existing retrieval-only approaches, both template retrieval and anomaly scoring rely on this similarity measure. However, directly searching across all template tokens in the database incurs substantial computational overhead and generalizes poorly to unseen categories. Moreover, when the number of templates is limited, as in few-shot scenarios, the detection performance becomes highly sensitive to noisy or unreliable templates.

Motivated by the RAG paradigm~\cite{RAGreviewLLM,RAGreview}, which alleviates hallucinations through augmented generation, we enhance UAD by further introducing a filtering-based generation reasoning process, as illustrated in Fig. \ref{pipeline}. The overall RAID framework can be formulated as:
\vspace{-1.5mm}
\begin{equation}\label{eq:RAG}
\mathcal{M} = \mathcal{G}(\{\textbf{p}_\mathcal{Q}\}, \mathcal{R}(\{\textbf{p}_\mathcal{Q}\},\mathcal{D})).
\vspace{-1.5mm}
\end{equation}
Here, $\mathcal{R}(\{\textbf{p}_\mathcal{Q}\},\mathcal{D})$ performs hierarchical retrieval from the database $\mathcal{D}$, balancing search efficiency and retrieval accuracy; $\mathcal{G}(\cdot,\cdot)$ then executes guided MoE filtering to generate the refined anomaly map $\mathcal{M}$, conditioned on the query and retrieved templates, effectively suppressing retrieval noise.

\subsection{Hierarchical Vector Database Construction}

Existing UAD methods typically adopt a flat retrieval structure~\cite{patchcore,guo2023template,Anomalydino}, where each query patch searches for its globally most similar counterpart in a large memory bank. Such a design leads to high computational overhead and degraded inference efficiency as the bank size grows. To balance retrieval accuracy and efficiency while ensuring inter-class discriminability and intra-class representational richness, we introduce a hierarchical vector database $\mathcal{D}$, as illustrated in Fig. \ref{pipeline}(a), with the following \texttt{indexing flow} of template tokens: class prototype $\{\bar{\mathbf{c}}\}\rightarrow$ semantic prototype $\{\bar{\mathbf{s}}\}\rightarrow$ instance token $\{\bar{\mathbf{t}}\}$. During database construction, both the CLS token $\mathbf{c}_\mathcal{T}$ and the patch tokens $\{\mathbf{p}_\mathcal{T}\}$ are utilized to form this multi-level hierarchy.

The \textbf{class prototype entity} indexes template tokens into corresponding class-level clusters. To obtain it, we perform K-means clustering on the CLS tokens from all templates:
\vspace{-1.5mm}
$$
\{\bar{\mathbf{c}}\} = \mathrm{KMeans}\bigl(\{\mathbf{c}_\mathcal{T}^{n}\}_{n=1}^{N}\bigr), 
\vspace{-1.5mm}
$$
where $\mathrm{KMeans}(\cdot)$ denotes the K-means operation, $N$ is the number of templates, and $\bar{\mathbf{c}}\in \mathbb{R}^{1\times D}$ represents the class prototype (class-level centroids), with $|\{\bar{\mathbf{c}}\}|=C$ as the class number. Leveraging the class-level semantics encoded in CLS tokens, the class prototype entity enables category- and dataset-agnostic retrieval, providing a strong scalability capability for multi-dataset UAD.

The \textbf{semantic prototype entity} organizes template tokens into an intra-class semantic structure:
\vspace{-1.5mm}
$$
\{\bar{\mathbf{s}}\}^c = \mathrm{KMeans}\bigl(\{\mathbf{p}_\mathcal{T}^{c,n}\}_{n=1}^{N^c}\bigr), 
\vspace{-1.5mm}
$$
where $\bar{\mathbf{s}}^c\in \mathbb{R}^{1\times D}$ represents the resulting semantic prototype. Note that multiple semantic prototypes are generated per class, and we use $\bar{\mathbf{s}}^{c,j}$ to indicate the $j$-th semantic prototype guided by the class prototype entity $\bar{\mathbf{c}}^c$. These prototypes capture recurring intra-class patterns, such as textures, structural components, or backgrounds, and serve as structured guidance for more effective anomaly detection.

The \textbf{instance token entity} stores all template tokens $\{\mathbf{p}_\mathcal{T}\}$ in organized instance token sets $\{\mathbf{t}\}^{c,j}$, sequentially indexed following the flow defined by the class prototype and semantic prototype entities. Here, we use $\mathbf{t}^{c,j,k}\in \mathbb{R}^{1\times D}$ to denote the $k$-th instance token associated with the semantic prototype $\bar{\mathbf{s}}^{c,j}$. This entity preserves fine-grained visual details, enabling accurate retrieval and pixel-level matching for downstream anomaly detection.


\subsection{Hierarchical Retrieval}
The hierarchical vector database naturally supports a coarse-to-fine \texttt{retrieval flow}, as demonstrated in Fig. \ref{pipeline}(a), which refines correspondences by narrowing the search space from global class-level clusters to intra-class semantics, and finally to local instance-level template tokens. This hierarchical retrieval mechanism ($\mathcal{R}$ in Eqn. (\ref{eq:RAG})) effectively reduces redundant matching and mitigates scalability bottlenecks observed in prior works~\cite{Anomalydino,CostFilterAD}, thereby enhancing the applicability of the proposed RAID framework to large-scale datasets.

At the top level, the CLS token of the input $\mathbf{c}_\mathcal{Q}$ is compared with the class prototypes $\{\bar{\mathbf{c}}\}$ via cosine similarity:
$$
\hat{c} = \arg\max\limits_{c} \mathrm{sim}(\mathbf{c}_\mathcal{Q}, \bar{\mathbf{c}}^c), \quad \bar{\mathbf{c}}^c \in \{\bar{\mathbf{c}}\}.
$$
The top-1 match provides an estimation of the input category $\hat{c}$. 

At the intermediate level, each patch token $\mathbf{p}_\mathcal{Q}$ from the input image queries the semantic prototype set $\{\bar{s}\}^{\hat{c}}$ of class $\hat{c}$, retrieving its top-$K'$ nearest semantic prototypes $\{\bar{\mathbf{s}}^{\hat{c},j}\}_{j=1}^{K'}$, defined as:
$$
\{\bar{\mathbf{s}}^{\hat{c},j}\}_{j=1}^{K'} = \arg\max\nolimits_{ \bar{\mathbf{s}} \in \{\bar{\mathbf{s}}\}^{\hat{c}} }^{K'} \mathrm{sim}(\mathbf{p}_\mathcal{Q}, \bar{\mathbf{s}}^{\hat{c},j}).
$$

Finally, at the lowest level, $\mathbf{p}_\mathcal{Q}$ further queries the instance token set $\{\mathbf{t}\}^{\hat{c},j}$ associated with its matched semantic prototypes $\{\bar{\mathbf{s}}^{\hat{c},j}\}_{j=1}^{K'}$, retrieving the top-$K$ most similar instance tokens $\{\mathbf{t}^{\hat{c},j,k}\}_{k=1}^K$, defined as:
$$
\{\mathbf{t}^{\hat{c},j,k}\}_{k=1}^K = \arg\max\nolimits_{\mathbf{t} \in \{\mathbf{t}\}^{\hat{c},j}}^K \ \mathrm{sim}(\mathbf{p}_\mathcal{Q}, \mathbf{t}^{\hat{c},j,k}).
$$

Finally, among the retrieved $K'$ semantic prototypes, only the most relevant one is retained for each patch token. Consequently, by efficiently traversing all patch tokens $\{\mathbf{p}_\mathcal{Q}\}$ of the input image, we prepare a total of $H'\times W'\times 1$ semantic prototypes and $H'\times W'\times K$ template tokens, each represented as a $1\times D$ feature vector.

\subsection{Filtering-based Generation Reasoning}
While the hierarchical retrieval flow effectively gathers template tokens, it also introduces hallucinatory noise from unreliable matches, spatial misalignment, and domain shifts. This noise blurs anomaly boundaries and obscures subtle defects. To address it, we reformulate the generation stage of RAG as a filtering-based generative reasoning process, employing a guided MoE filter that adaptively denoises and refines the anomaly cost volume, as shown in Fig. \ref{pipeline}(b).

\textbf{Initial anomaly cost volume.} For each query token $\mathbf{p}_\mathcal{Q}^{(y,x)}$ at spatial coordinate $(y,x)$, we define patch-level anomaly cost with its retrieved instance patches $\mathbf{t}^{(y,x),k}\in \{\mathbf{t}^{(y,x),k}\}_{k=1}^K$ based on cosine similarity in Eqn. (\ref{eq:cos_sim}):
\vspace{-1.5mm}
$$
\mathcal{C}^{y,x,k} = 1-\mathrm{sim}(\mathbf{p}_\mathcal{Q}^{(y,x)}, \mathbf{t}^{(y,x),k}),
\vspace{-1.5mm}
$$
where $\mathcal{C}\in\mathbb{R}^{H'\times W'\times K}$ is the resulting 3D anomaly cost volume, with $(y,x)\in\mathbb{R}^{H'\times W'}$ indicating spatial positions and $k\in\mathbb{R}^K$ indexing the matching candidates. Note that lower similarity values correspond to higher anomaly likelihoods.

By leveraging the hierarchical retrieval mechanism, RAID selects only a small set of highly relevant candidates for each query token, resulting in a compact and well-aligned cost volume. This design contrasts with CostFilter-AD~\cite{CostFilterAD}, which constructs a global matching space, thereby significantly improving inference efficiency.

\textbf{Guided MoE filtering.} The guided MoE filter is designed as a two-stage architecture: the first stage constructs a guidance map via dual-guidance fusion, while the second stage performs guided filtering.

In the first stage, the semantic prototypes $\{\bar{\mathbf{s}}\}$ are rearranged into image-like prototype guidance maps $g_s\in\mathbb{R}^{H'\times W'\times D\times 1}$, and the query tokens $\{\mathbf{p}_\mathcal{Q}\}$ are rearranged into image-like query guidance maps $g_\mathcal{Q}\in\mathbb{R}^{H'\times W'\times D\times K}$. A convolutional router takes the concatenation $\text{cat}(g_\mathcal{Q}, g_s)$ as input to compute sparse routing probabilities and weights, then aggregates the activated experts to produce the fused guidance $\tilde{g}$:
$$
\begin{aligned}
    p &= \text{Softmax}(\text{Router}(\text{cat}(g_\mathcal{Q}, g_s))),\\
    \tilde{p}^{i} &= 
    \begin{cases}
    p^{i}, & i \in \mathrm{Top}\text{-}k(p), \\
    0, & \text{otherwise},\\
    \end{cases}\\
    \tilde{g} &= \sum\nolimits_{i=1}^M \tilde{p}^i E^i_{g}\left(\text{cat}(g_\mathcal{Q}, g_s)\right),
\end{aligned}
$$
where $E^i_{g}$ denotes the $i$-th convolutional expert used for guidance fusion, and $\tilde{p}^i$ is its routing weight. Only the top-$k$ experts with the highest scores are activated, encouraging specialization across distinct semantic patterns.

In the second stage, the initial anomaly cost volume $\mathcal{C}$ is refined via denoising MoE, where a router softly activates all experts. Each denoising expert $E_\mathcal{C}^i$ performs dual-branch filtering of $\mathcal{C}$ under the fused guidance $\tilde{g}$, which consists of a cross-attention branch ($\tilde{g}$ as query and $\mathcal{C}$ as key/value) and a convolutional branch. The router assigns dense expert weights $p^i$ and aggregates the expert outputs $\tilde{\mathcal{C}}^i$ to yield the final anomaly map:
$$
\mathcal{M} = \sum^M\nolimits_{i=1}p^i\cdot \tilde{\mathcal{C}}^i, \ \ \tilde{\mathcal{C}}^i=E_\mathcal{C}^i(\tilde{g},\mathcal{C}),$$
where $\mathcal{M}$ denotes the generated anomaly map. The complete MoE architecture is detailed in the Appendix.

\begin{table*}[t]
\renewcommand{\arraystretch}{0.85}
    \centering
    \caption{\textbf{Full-shot} (multi-class UAD) performance (\%) on four industrial datasets. Best results are highlighted in \textbf{bold}.}
    \label{tab:tab_mvtec}
    \vspace{-2mm}
    \footnotesize
\setlength{\tabcolsep}{11pt}
    \begin{tabular}{ll c c c c c c c c}
        \toprule
        \multirow{2}{*}{Dataset} & \multirow{2}{*}{Method} &  & \multicolumn{3}{c}{Image-level} & \multicolumn{4}{c}{Pixel-level} \\
\cmidrule(lr){4-6} \cmidrule(lr){7-10}
&  &  & AUROC & AP & F1-max & AUROC & AP & F1-max & AUPRO \\
\midrule
 \multirow{10}{*}{MVTec-AD~\cite{mvtec}} & PatchCore \cite{patchcore} & & 96.4 &-&-& 95.7&-&-&-\\
 & UniAD \cite{uniad} && 96.5 &98.8 &96.2 &96.8 &43.4 &49.5 &90.7\\
 & SimpleNet \cite{simplenet} & &95.3 &98.4 &95.8 &96.9 &45.9 &49.7 &86.5\\
 & MambaAD \cite{mambaad} & & 98.6&99.6&97.8	&97.7&56.3&59.2&93.1
\\
 & GLAD \cite{glad} & &97.5&98.8&96.8&	97.3&58.8&59.7&92.8
\\
& DiAD \cite{diad} & & 97.2 &99.0&96.5&	96.8&52.6&55.5&90.7\\
 & ViTAD \cite{vitad} & & 98.3&99.4&97.3	&97.7&55.3&58.7&91.4
\\
 & AnomalyDINO \cite{Anomalydino} & & 96.8&98.6&97.1&	98.1&61.3&60.8&93.6
\\
 & Costfilter-AD \cite{CostFilterAD} && 99.0&99.7&98.6&98.0&58.1&61.2&93.2
 \\
 & RAID (Ours)  & & \textbf{99.4} &\textbf{99.8}	&\textbf{98.7}&\textbf{98.6}&	\textbf{71.7}&	\textbf{68.5}&	\textbf{95.6}
\\
    \midrule
    
\multirow{9}{*}{VisA \cite{visa}} 
 & UniAD \cite{uniad} && 85.5 &85.5 &84.4 &95.9 &21.0 &27.0 &75.6\\
 & SimpleNet \cite{simplenet} & &87.2 &87.0 &81.8 &96.8 &34.7 &37.8 &81.4\\
 & MambaAD \cite{mambaad} & & 94.3 &94.5 &89.4 &98.5 &39.4 &44.0 &91.0
\\
 & GLAD \cite{glad} & &90.1&91.4&86.7	&97.4&33.9&39.4&91.5

\\
& DiAD \cite{diad} & & 86.8 &88.3 &85.1 &96.0 &26.1 &33.0 &75.2\\
 & ViTAD \cite{vitad} & & 90.5 &91.7 &86.3 &98.2 &36.6 &41.1 &85.1
\\
 & AnomalyDINO \cite{Anomalydino} & & 90.5&91.4&86.2	&97.5&39.6&40.4&86.3

\\
 & Costfilter-AD \cite{CostFilterAD} && 93.4& 95.2&89.3	&98.6&41.4&45.0&86.8

 \\
 & RAID (Ours) & & \textbf{94.9}	& \textbf{95.5}	&\textbf{90.6}	&\textbf{99.0}	&\textbf{45.2}	&\textbf{49.2}	&\textbf{91.7}
\\
    \midrule
    
\multirow{10}{*}{MPDD \cite{mpdd}} & PatchCore \cite{patchcore}& &83.5& -& -&	97.7&-&-&-
\\
& UniAD \cite{uniad}& &80.1& 83.2& 85.1 &95.4 &19.0 &25.6 &83.8\\
 & Hvq-Trans \cite{uniad} && 86.5&87.9&85.6&	96.9&26.4&30.5&88.0
\\
& SimpleNet \cite{simplenet} &&90.6& 94.1 &89.7 &97.1 &33.6& 35.7& 90.0 \\
& MambaAD \cite{mambaad} &&89.2& 93.1& 90.3& 97.7& 33.5& 38.6 &92.8\\
 & GLAD \cite{glad} & &90.8&90.5&90.2&	98.0&40.0&40.6&93.1
\\

& DiAD \cite{diad} & & 85.8&89.2&86.5	&91.4&15.3&19.2&66.1
\\
 & ViTAD \cite{vitad} & & 87.4&90.8&87.0	&97.8&44.1&46.4&95.3
\\
 & Costfilter-AD \cite{CostFilterAD} && 93.1&95.4&90.3	&97.5&34.1&37.0&82.9
 \\
 & RAID (Ours) &&\textbf{96.3}&\textbf{97.6}&\textbf{95.0}	&\textbf{98.9}&\textbf{47.0}&\textbf{46.9}&\textbf{96.6}
\\
    \midrule
    
\multirow{8}{*}{BTAD \cite{btad}}  
& UniAD \cite{uniad}&& 94.5 &98.4 &94.9& 97.4& 52.4& 55.5& \textbf{78.9}\\
 & Hvq-Trans \cite{uniad} && 90.9 & 97.8 & 94.8	&96.7 & 43.2 &48.7 & 75.6
\\
& SimpleNet \cite{simplenet} &&94.0& 97.9& 93.9& 96.2& 41.0& 43.7& 69.6\\
& MambaAD \cite{mambaad}& &92.9 &96.2& 93.0& \textbf{97.6}& 51.2& 55.1& 77.3\\
& DiAD \cite{diad} & & 90.2 &88.3 &92.6 	&91.9 &20.5 &27.0 &70.3

\\
 & ViTAD \cite{vitad} & & 94.0 &97.0 &93.7 	&\textbf{97.6} &58.3 &56.5 &72.8
\\
 & Costfilter-AD \cite{CostFilterAD} && 93.3 &\textbf{98.6} &\textbf{96.0}&97.3&47.0&50.2&76.2
 \\
 & RAID (Ours) &&\textbf{95.2}&96.3&93.0	&\textbf{97.6}&\textbf{67.3}&\textbf{64.3}&72.2

\\
    \midrule
    \end{tabular}
    \vspace{-4mm}
\end{table*}

\subsection{Training and Inference}
We adopt a self-supervised training strategy commonly used in UAD~\cite{draem,glad,08}, where synthetic anomalous images are paired with their corresponding synthetic anomaly masks $\mathcal{M}_s$. The overall objective function is defined as:
$$
\mathcal{L} = \mathcal{L}_{\text{focal}}(\mathcal{M},\mathcal{M}_s) + \lambda_\text{bal} \mathcal{L}_{\text{bal}},
$$
where the focal loss \cite{Focalloss}$ ~\mathcal{L}_{\text{focal}}$ addresses the inherent imbalance between normal and anomalous pixels, $\mathcal{L}_{\text{bal}}$ regularizes the expert routing process to prevent bias toward dominant experts and mitigate router collapse~\cite{fedus2022switch}, and $\lambda_\text{bal}$ is the balancing weight. 

For inference, the anomaly cost volume is constructed and filtered to yield the refined anomaly map $\mathcal{M}$. The image-level anomaly score is computed as the mean of the top 1\% highest responses in $\mathcal{M}$, while for pixel-level localization, $\mathcal{M}$ is directly used as the anomaly map.

\section{Experiments}
\subsection{Experimental Settings}
\textbf{Datasets.} We evaluate our method on four widely used industrial anomaly detection benchmarks, MVTec-AD \cite{mvtec}, VisA \cite{visa}, MPDD \cite{mpdd}, and BTAD \cite{btad}, which cover diverse anomaly types, object complexities, and imaging conditions, offering a comprehensive benchmark for assessing UAD robustness and generalization.
\textbf{MVTec-AD} contains 5,354 high-resolution images of 10 objects and 5 textures, with normal samples for training and diverse defects for testing. \textbf{VisA} includes 10,821 images across 12 object subsets, covering various surface and structural anomalies such as scratches, dents, and color spots. \textbf{MPDD} focuses on metallic parts, providing 1,346 images under challenging backgrounds and illumination. \textbf{BTAD} comprises 2,830 images from three industrial categories, featuring both normal and defective samples with real-world variability. 

\textbf{Evaluation metrics.} 
Following common practice, we primarily report the Area Under the Receiver Operating Characteristic Curve for image-level anomaly detection (I-AUROC) and pixel-level anomaly localization (P-AUROC). Additional metrics, including Average Precision (AP), the maximum F1 score (F1-max), and the pixel-level Area Under the Per-Region Overlap (AUPRO), are also evaluated.

\textbf{Implementation details.} For full-shot and multi-dataset experiments, all input images are resized to $256 \times256$. We use DINOv2-s \cite{dinov2} as the feature extractor. In constructing the hierarchical vector database, 80 templates are used for MVTec-AD and 100 for VisA, while all normal samples serve as templates for MPDD and BTAD. Patch tokens within each class are clustered into 50 semantic prototypes. During retrieval, the $K'=5$ nearest semantic prototypes and $K=150$ instance tokens are retrieved for each query token. The guided MoE filter in the generation stage includes three experts per layer, with sparse routing in the first layer activating two experts per input. The weighting parameter $\lambda_\text{bal}$ is 0.005. Models are trained for 100 epochs using Adam with an initial learning rate of $1\times10^{-4}$. For few-shot evaluation, the same settings are applied except that input images are resized to $224\times224$ and the number of templates is adjusted accordingly.

\subsection{Quantitative Comparison}
We comprehensively evaluate the effectiveness and generalization of RAID under the multi-class UAD paradigm. The extensive experiments are conducted on four widely used benchmarks: MVTec-AD \cite{mvtec}, VisA \cite{visa}, MPDD \cite{mpdd}, and BTAD \cite{btad}, under three representative settings: \textbf{full-shot}, \textbf{few-shot}, and \textbf{multi-dataset}. For the \textbf{full-shot} scenario, we adopt the ``one-model-for-all classes'' paradigm and compare our method with a diverse set of State-Of-The-Art (SOTA) baselines, including PatchCore \cite{patchcore}, HVQ-Trans \cite{HVQTrans}, GLAD \cite{glad}, DiAD \cite{diad}, ViTAD \cite{vitad}, CostFilter-AD \cite{CostFilterAD}, UniAD \cite{uniad}, SimpleNet  \cite{simplenet}, MambaAD \cite{mambaad}, and AnomalyDINO (full-shot) \cite{Anomalydino}, covering feature-embedding, diffusion-based, and transformer-based frameworks. For the \textbf{few-shot} scenario, we evaluate against PatchCore \cite{patchcore}, Win-CLIP \cite{jeong2023winclip}, FastRecon \cite{fang2023fastrecon}, PromptAD \cite{promptad}, IIPAD \cite{iipad}, and DFM \cite{dfm}, to assess the model performance using limited normal samples \textbf{without any fine-tuning}. For the \textbf{multi-dataset} scenario, we adopt the ``one-model-for-all datasets'' paradigm, jointly training on multiple datasets and comparing it with the representative OneNIP \cite{onenip} to validate cross-dataset generalization.

\begin{table*}[t]
    \centering
\renewcommand{\arraystretch}{0.9}
    \tabcolsep=0.15cm
    \caption{\textbf{Few-shot} performance comparison under the \textbf{multi-class} paradigm on MVTec-AD and VisA using I-AUROC/P-AUROC.}
    \label{tab:fewshot}
    \vspace{-2mm}
    \footnotesize
\setlength{\tabcolsep}{7pt}
    \begin{tabular}{cc|cccccc|c}
        \toprule
     Setup & Datasets & PatchCore \cite{patchcore} & Win-CLIP \cite{jeong2023winclip} & FastRecon \cite{fang2023fastrecon} & PromptAD \cite{promptad} & IIPAD \cite{iipad} & DFM \cite{dfm} & RAID (Ours) \\
      \midrule
      \multirow{2}{*}{1-shot}& MVTec-AD & 86.3 / 93.3 &  92.6 / 91.6 & 83.7 / 93.9 & 93.0 / 95.2 & 94.2 / 96.4 & 87.2 / 95.2 & \textbf{95.1} / \textbf{96.6} \\ 
      & VisA & 79.9 / 95.4 &  84.8 / 95.3 & 80.1 / 96.5 & 85.2	/ 97.2 & 85.4 / 96.9 & 84.0 / 96.4 & \textbf{85.8} / \textbf{97.7}\\
      \midrule
       \multirow{2}{*}{2-shot}& MVTec-AD & 83.4 / 92.0  & 93.8 / 91.9 & 88.9 / 95.3 & 95.4 / 95.6 & 95.7 / 96.7 & 90.2 / 96.0 & \textbf{96.6} / \textbf{97.1}\\
       & VisA & 81.6 / 96.1 & 83.5 / 95.7 & 84.6 / 97.5 & 85.1	/ 97.7 & 86.7	/ 97.2 & 86.0 / 96.8 & \textbf{88.5} / \textbf{97.9}\\
       \midrule
       \multirow{2}{*}{4-shot}& MVTec-AD & 88.8 / 94.3 & 95.5 / 92.4 & 94.2 / 95.9 & 95.9 / 96.0 &96.1 / 97.0 &	92.0 / 96.2 & \textbf{96.9} / \textbf{96.9}\\
       &VisA & 85.3 / 96.8 & 85.7 / 96.0 & 68.5 / 96.0 & 87.5 / 97.9
 & 88.3	/ 97.4 & \textbf{89.8} / 97.1 & 89.3 / \textbf{98.2}\\ 
        \bottomrule
    \end{tabular}
    \vspace{-4mm}
\end{table*}

\begin{table}[t]
    \centering
    \footnotesize
    \tabcolsep=0.12cm
    \caption{\textbf{Multi-dataset} performance comparison of \textbf{a single model} jointly trained on MVTec-AD, VisA, MPDD, and BTAD using I-AUROC/I-AP/P-AUROC/P-AP.
    }
    \vspace{-2mm}
    \label{tab:tab_multi}
    \begin{tabular}{c|c|c|c}
        \toprule
      Datasets & \# class & OneNIP \cite{onenip}  & RAID (Ours) \\
      \midrule
     \scriptsize{MVTec-AD} & 15 &  96.7 / 98.8 / 97.1 / 57.5 & 99.4 / 99.8 / 98.5 / 69.4\\ 
      VisA & 12 &  92.8 /	95.1 / 98.7	/ 41.1 & 94.6 / 94.8 / 98.8 / 47.1\\
      MPDD & 6 & 85.2 / 88.4 / 97.8 / 37.0 & 93.6 / 96.1 / 98.8 / 44.9\\
      BTAD & 3 & 93.2 / 96.5 / 97.9 / 59.8 & 93.8 / 96.0 / 97.8 / 66.4\\
      \midrule
      All & 36 & 92.0 / 94.7 / 97.9 / 48.9 & \textbf{95.4} / \textbf{96.7} / \textbf{98.5} / \textbf{57.0}  \\
        \bottomrule
    \end{tabular}
    \vspace{-4mm}
\end{table}

\textbf{Full-shot for multi-class UAD.}
We evaluate image\mbox{-}level detection and pixel\mbox{-}level localization for full\mbox{-}shot UAD, where the training phase has access to all normal samples on four benchmarks, as shown in Table~\ref{tab:tab_mvtec}. On \textbf{MVTec\mbox{-}AD} \cite{mvtec}, our method \emph{outperforms} GLAD \cite{glad}, AnomalyDINO \cite{Anomalydino}, and CostFilter\mbox{-}AD \cite{CostFilterAD} by \(1.9\% / 1.3\%\), \(1.6\% / 0.5\%\), and \(0.4\% / 0.6\%\) in I\mbox{-}AUROC/P\mbox{-}AUROC, respectively. On \textbf{VisA} \cite{visa}, it reaches SOTA with gains of \(4.8\% / 1.6\%\), \(4.4\% / 1.5\%\), and \(1.5\% / 0.4\%\). On \textbf{MPDD} \cite{mpdd}, advances are \(5.5\%\), \(8.9\%\), and \(3.2\%\) for I\mbox{-}AUROC and \(0.9\%\), \(1.1\%\), and \(1.4\%\) for P\mbox{-}AUROC over GLAD, ViTAD, and CostFilter\mbox{-}AD. On \textbf{BTAD} \cite{btad}, it exceeds CostFilter\mbox{-}AD and ViTAD by \(1.9\%\) and \(1.2\%\) in I\mbox{-}AUROC. In addition, our method also delivers strong performance on other metrics involving image\mbox{-}level AP/F1\mbox{-}max, and pixel\mbox{-}level AP/F1\mbox{-}max/AUPRO. Collectively, the consistent gains across datasets support our RAG\mbox{-}inspired design: hierarchical retrieval supplies contextually relevant references, and the guided filter refines the anomaly cost volume into reliable image\mbox{-}level decisions and precise pixel\mbox{-}level localization.

\textbf{Few-shot UAD generalizability.}
We evaluate RAID in the few-shot setting, resizing inputs to \(224\times224\) for a fair comparison with DFM~\cite{iipad}, IIPAD~\cite{dfm}, and PromptAD~\cite{dfm}. Following common practice, the model is trained on the auxiliary MPDD dataset and then tested on MVTec-AD~\cite{mvtec} and VisA~\cite{visa}. Table~\ref{tab:fewshot} reports averaged results over five seeds, RAID attains consistently high I-AUROC/P-AUROC (standard deviations in the Appendix). On MVTec-AD, it surpasses DFM~\cite{dfm} by \(7.9\% / 1.4\%\), \(6.4\% / 1.1\%\), and \(4.9\% / 0.7\%\) for the 1-, 2-, and 4-shot settings (I-AUROC/P-AUROC).
Notably, RAID operates in an image-only manner, unlike PromptAD~\cite{promptad} and WinCLIP~\cite{jeong2023winclip} that leverage language priors, and it uses a lower input resolution than FastRecon~\cite{fang2023fastrecon} and WinCLIP~\cite{jeong2023winclip}. These results highlight the strong generalization capability of RAID, which redefines UAD through the RAG perspective: contextually relevant references retrieved from the database iteratively refine the anomaly cost volume and guide generation toward semantically coherent regions, leading to more accurate and transferable anomaly localization.

\begin{figure}
\begin{center}
\centerline{\includegraphics[width=0.999\linewidth]{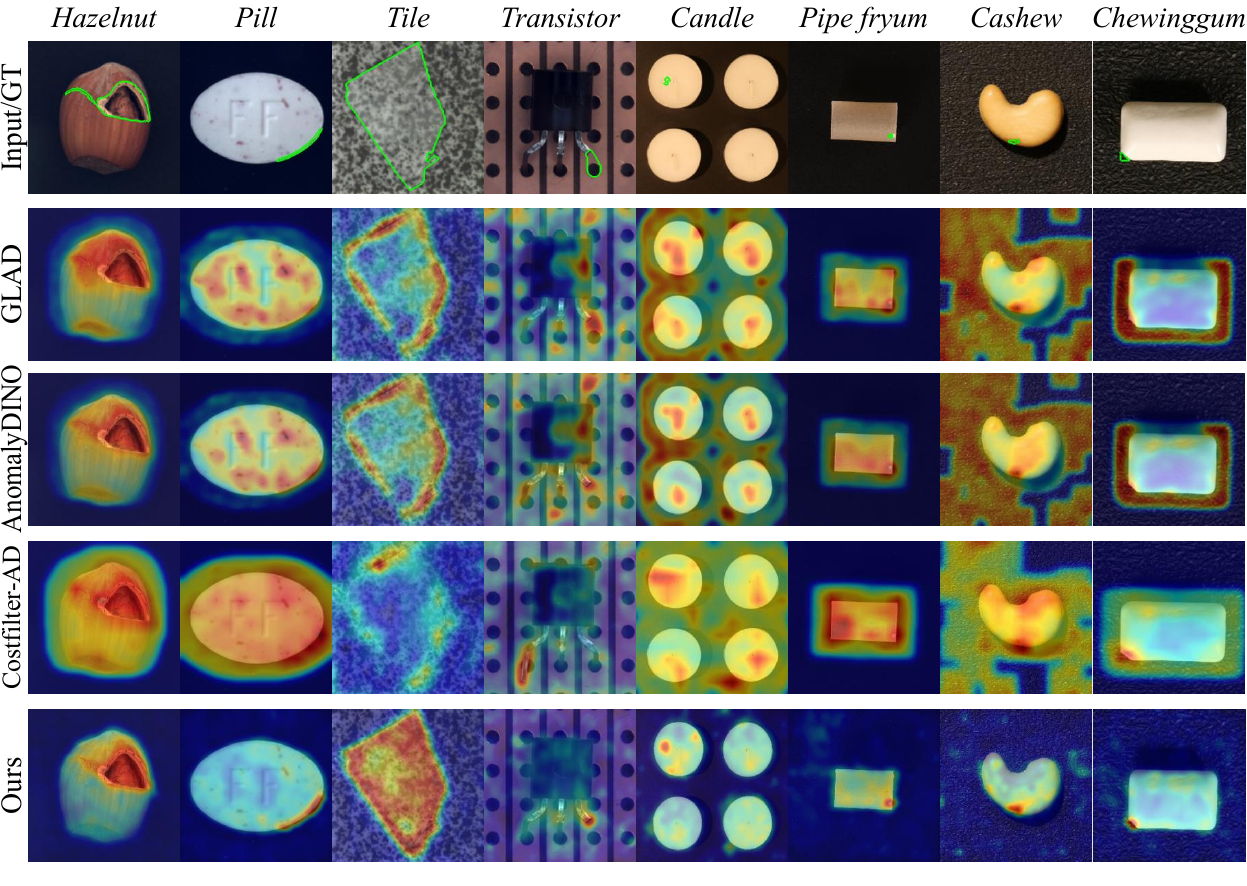}}
\vspace{-2mm}
\caption{Qualitative comparison of multi-class anomaly localization results on MVTec-AD and VisA datasets.}
\label{fig:Qualitative}
\end{center}
\vspace{-8mm}
\end{figure}


\textbf{Multi-dataset UAD scalability.}
We merge MVTec-AD~\cite{mvtec}, VisA~\cite{visa}, BTAD~\cite{btad}, and MPDD~\cite{mpdd} into a larger and more diverse dataset following OneNIP~\cite{onenip}, and train RAID on this merged corpus.
Table~\ref{tab:tab_multi} reports image-level classification and pixel-level segmentation, including the overall average across \(36\) categories and the per dataset averages.
In this multi-dataset setting, RAID surpasses OneNIP~\cite{onenip} in image-level classification \((95.4\% / 96.7\%)\) and pixel-level segmentation \((98.5\% / 57.0\%)\), whereas OneNIP attains \(92.0\% / 94.7\% / 97.9\% / 48.9\%\) on the corresponding metrics.
Moreover, moving from the multi-class setting in Table~\ref{tab:tab_mvtec} to the unified regime of the dataset across Table~\ref{tab:tab_multi}, our method shows no noticeable degradation, thus establishing the proposed RAID as a scalable and effective solution for UAD in complex distributions.

\subsection{Qualitative Comparison}
We present qualitative comparisons on the MVTec-AD~\cite{mvtec} and VisA~\cite{visa} benchmarks to further validate the effectiveness of our method. As shown in Fig.~\ref{fig:Qualitative}, our approach outperforms GLAD~\cite{glad}, AnomalyDINO~\cite{Anomalydino}, and CostFilter-AD~\cite{CostFilterAD}, achieving more precise anomaly localization with sharper boundaries, reduced matching noise, and enhanced sensitivity to subtle defects, owing to the RAG-like paradigm. Please refer to the Appendix for more visualization results.


\subsection{Ablation Studies and Further Analysis}

\begin{table}[t]
    \centering
    \footnotesize
    \tabcolsep=0.06cm
    \caption{The efficiency and accuracy (I\mbox{-}AUROC and P\mbox{-}AUROC) comparison on MVTec-AD under the different retrieval and database (DB.) capacity settings. }
    \label{tab:tab_efficiency}
    \vspace{-2mm}
    \begin{tabular}{c|c|c|c|c|c|c}
        \toprule
      Methods &  Retrieval & DB. & FLOPs & Mem. (GB)  & Inf. (s)& Results \\
      \midrule
     PatchCore & Flat & Multi&7.12G&3.46&0.093&96.4 / 95.7\\
     PatchCore & Flat & Single &7.12G &2.48&0.054&99.0 / 98.0\\
     \scriptsize{AnomalyDINO} & Flat & Single&4.90G&3.25 & 0.067 & 96.8 / 98.1\\ 
     
     RAID & Flat & Multi & 14.2G & 6.55 & 0.267 &99.4 / 98.7\\
     RAID & Hier. & Single & 14.2G &6.52&0.052&99.3 / 98.7 \\
     RAID & Hier.& Multi & 14.2G & 6.52 & 0.062 & 99.4 / 98.6 \\
     RAID & Hier.& 4-shot & 14.2G  &  5.28 & 0.046 & 96.9 / 96.9 \\ 
        \bottomrule
    \end{tabular}
    \vspace{-3mm}
\end{table}

\textbf{Effectiveness of the retrieval strategy.}
In Table~\ref{tab:tab_efficiency}, we compare two retrieval schemes (flat vs. hierarchical) under three database capacity settings (multi\mbox{-}class, single\mbox{-}class, and few\mbox{-}shot). In practice, the hierarchical retrieval achieves about \(5\times\) lower per-image latency than the flat scheme \(0.267s\) (which prioritizes accuracy over speed) while maintaining nearly identical I\mbox{-}AUROC and P\mbox{-}AUROC of \(99.4\% / 98.6\%\). 
This demonstrates the efficiency–precision balance of our hierarchical retrieval scheme. Across the database capacities, multi\mbox{-}class and single\mbox{-}class settings yield comparable accuracy (\(99.4\% / 98.6\%\) vs. \(99.3\% / 98.7\%\)), and even a 4\mbox{-}shot setup preserves strong performance \(96.9\% / 96.9\%\) at a similar per-image runtime (\(0.046s\)), highlighting the efficient use of limited normal templates. Overall, these observations validate the design of the hierarchical vector database and the associated retrieval strategy.

\begin{table}[t]
    \centering
    \footnotesize
    \tabcolsep=0.08cm
    \caption{Effectiveness of template quantity on MVTec-AD.}
    \label{tab:tab_visA_pixel}
    \vspace{-2mm}
    \begin{tabular}{c|c|c|c|c|c}
        \toprule
      \# template & 20 & 40  & 60 & 80 & All \\
      \midrule
     Metrics & 98.1 / 97.7 & 98.7 / 97.8 & 98.9 / 98.1 & \textbf{99.4} / \textbf{98.6} & 99.3 / \textbf{98.6}\\ 
        \bottomrule
    \end{tabular}
    \vspace{-3mm}
\end{table}

\textbf{Analysis of template quantity.}
We vary the per-class template count on MVTec\mbox{-}AD from 20 to \emph{All}, as shown in Table~\ref{tab:tab_visA_pixel}. The performance steadily improves from \(98.1\% / 97.7\%\) (I\mbox{-}AUROC/P\mbox{-}AUROC) to a peak of \(99.4\% / 98.6\%\) at 80 templates, while the \emph{All} setting remains near-saturated at \(99.3\% / 98.6\%\). The gain up to 80 templates validates the retrieval pipeline: expanding the relevant template pool broadens normal-mode coverage and enhances query–template contrast, while hierarchical retrieval compacts matching candidates into a cleaner cost volume for the guided MoE to refine. The near-saturation under \emph{All} suggests that, while further increasing the template count may offer marginal gains, relevance remains the dominant factor in achieving strong detection and localization with a compact, well-curated template database.

\textbf{Effectiveness of the guided MoE filter.}
Table~\ref{tab:tab_ablation_bi} analyzes how the guided MoE filter transforms the retrieval-built cost volume into reliable anomaly decisions. Starting from retrieval only (ID0, \(97.9\% / 97.5\%\)), adding the cross-attention branch (Cro-Att.) and the Router\(_\mathcal{C}\) in the 2nd MoE stage (ID1) raises accuracy to \(98.5\% / 97.6\%\) by  refining semantically consistent matches. Incorporating the first-stage MoE\(_g\) (ID2) further boosts accuracy to \(99.2\% / 98.4\%\), demonstrating the benefits of dual-guidance fusion. The convolutional branch (Conv.) in the second-stage (ID3) refines local responses (\(98.7\% / 97.8\%\)), while its combination with cross-attention (ID4, \(99.1\% / 98.1\%\)) or MoE\(_g\) (ID5, \(98.9\% / 98.2\%\)) brings complementary gains. The former strengthens denoising with a global perception, and the latter refines the guidance. Removing Router\(_\mathcal{C}\) (ID6) leads to a drop (\(98.0\% / 97.5\%\)), confirming the need for sparse routing to preserve expert specialization. The full configuration (ID7) achieves the best (\(99.4\% / 98.6\%\)), validating that the guided MoE filter jointly denoises and refines the cost volume, delivering consistent improvements in both detection and localization.

\begin{table}[t]
    \centering
\renewcommand{\arraystretch}{0.9}
    \caption{Ablation studies of guided MoE filter on MVTec-AD.}
    \label{tab:tab_ablation_bi}
    \vspace{-2mm}
    \footnotesize
     \begin{tabular}{c|cccc|c}
        \toprule
        ID & MoE$_g$ & Cro-Att. & Conv. & Router$_\mathcal{C}$ & I\mbox{-}/P\mbox{-}AUROC \\
        \midrule
        0 & - & - & - & - & 97.9 / 97.5\\
        1 & - & \checkmark & - & \checkmark & 98.5 / 97.6 \\
        2 & \checkmark & \checkmark & - & \checkmark & 99.2 / 98.4\\
        3 & - & - & \checkmark & \checkmark & 98.7 / 97.8  \\
        4 & - & \checkmark & \checkmark & \checkmark & 99.1 / 98.1 \\
        5 & \checkmark & - & \checkmark & \checkmark & 98.9 / 98.2 \\
        6 & - & \checkmark & \checkmark & - & 98.0 / 97.5\\
        7 & \checkmark & \checkmark & \checkmark & \checkmark & \textbf{99.4} / \textbf{98.6}\\
        \bottomrule
    \end{tabular}
    \vspace{-3mm}
\end{table}

\begin{table}[t]
    \centering
    \footnotesize
    \tabcolsep=0.05cm
    \caption{Effectiveness of expert quantity on MVTec-AD.}
    \vspace{-2mm}
    \label{tab:tab_expert_quantity}
    \begin{tabular}{c|c|c|c|c|cc}
        \toprule 
      \text{\# ($E_g, E_\mathcal{C}$)}  & (3, 1) & (3, 2) & (2, 3) & (3, 3) & (4, 3) \\
      
      \midrule
     Metrics & 99.0 / 98.2 & 99.2 / 98.4 & 99.0 / 98.1 & \textbf{99.4} / \textbf{98.6} & 98.8 / 97.9\\

        \bottomrule
    \end{tabular}
    \vspace{-3mm}
\end{table}

\textbf{Expert quantity in MoE.}
Table~\ref{tab:tab_expert_quantity} analyzes the impact of expert numbers in the two-stage guided MoE filtering \((E_g, E_\mathcal{C})\). Performance peaks at \((3,3)\) with \(99.4\% / 98.6\%\) (I-AUROC/P-AUROC), indicating optimal specialization. Increasing the denoising experts $E_\mathcal{C}$ from 1 to 3 improves accuracy, while reducing guidance diversity \(E_g=2\) or over-expanding experts  \(E_g=4\) degrades results due to reduced specialization and diluted guidance. These results highlight a capacity-specialization tradeoff, with $(3,3)$ achieving the best balance for reliable anomaly detection.


\section{Conclusion} 
We presented RAID, a Retrieval-Augmented Industrial anomaly Detection framework that revisits UAD from a RAG perspective. By integrating hierarchical retrieval with guided MoE filtering-based generation, RAID effectively suppresses matching noise and preserves fine-grained anomaly boundaries and subtle anomalies. 
Extensive experiments across full-shot, few-shot, and multi-dataset settings on four benchmarks demonstrate that RAID consistently outperforms prior SOTA methods, achieving both robust generalization and scalability.
Looking forward, we envision that bringing the RAG paradigm into agentic and cross-modal anomaly detection opens a new direction toward more explainable, scalable, and data-efficient industrial intelligence.

\section*{Acknowledgements}
This work is supported in part by the Natural Science Foundation of Liaoning Province of China under Grant No. 2024-MSBA-42, in part by the Key Research and Development Program of Liaoning Province under Grant No. 2023JH26/10200011, and in part by the Science and Technology Major Project of Liaoning Province under Grant No. 2024JH1/11700048.
{
    \small
    \bibliographystyle{ieeenat_fullname}
    \bibliography{main}
}

\maketitlesupplementary

\setcounter{section}{0}
\renewcommand{\thesection}{S\arabic{section}}
In this supplementary material, we present the detailed architecture of the guided MoE filter (Sec.~\ref{supp:moe}), a feasibility analysis of category- and dataset-agnostic retrieval (Sec.~\ref{supp:retrieval}), additional results \textbf{applying RAID to reconstruction-based approaches} (Sec.~\ref{supp:recon}), ablation studies (Sec. \ref{sec:ablation_retrieval}, \ref{sec:ablation_study_visa}), Visualization of retrieval-reasoning interaction (Sec. \ref{sec:retrieval-reasoning}), the analysis of expert specialization (Sec. \ref{sec:expert_specialization}), extended quantitative and visualization results (Sec.~\ref{supp:results1}-\ref{supp:results3}), and an analysis of representative failure cases (Sec.~\ref{supp:fail}).

\section{Detailed Architecture of MoE Filter}\label{supp:moe}
\begin{figure}
\begin{center}
\centerline{\includegraphics[width=0.999\linewidth]{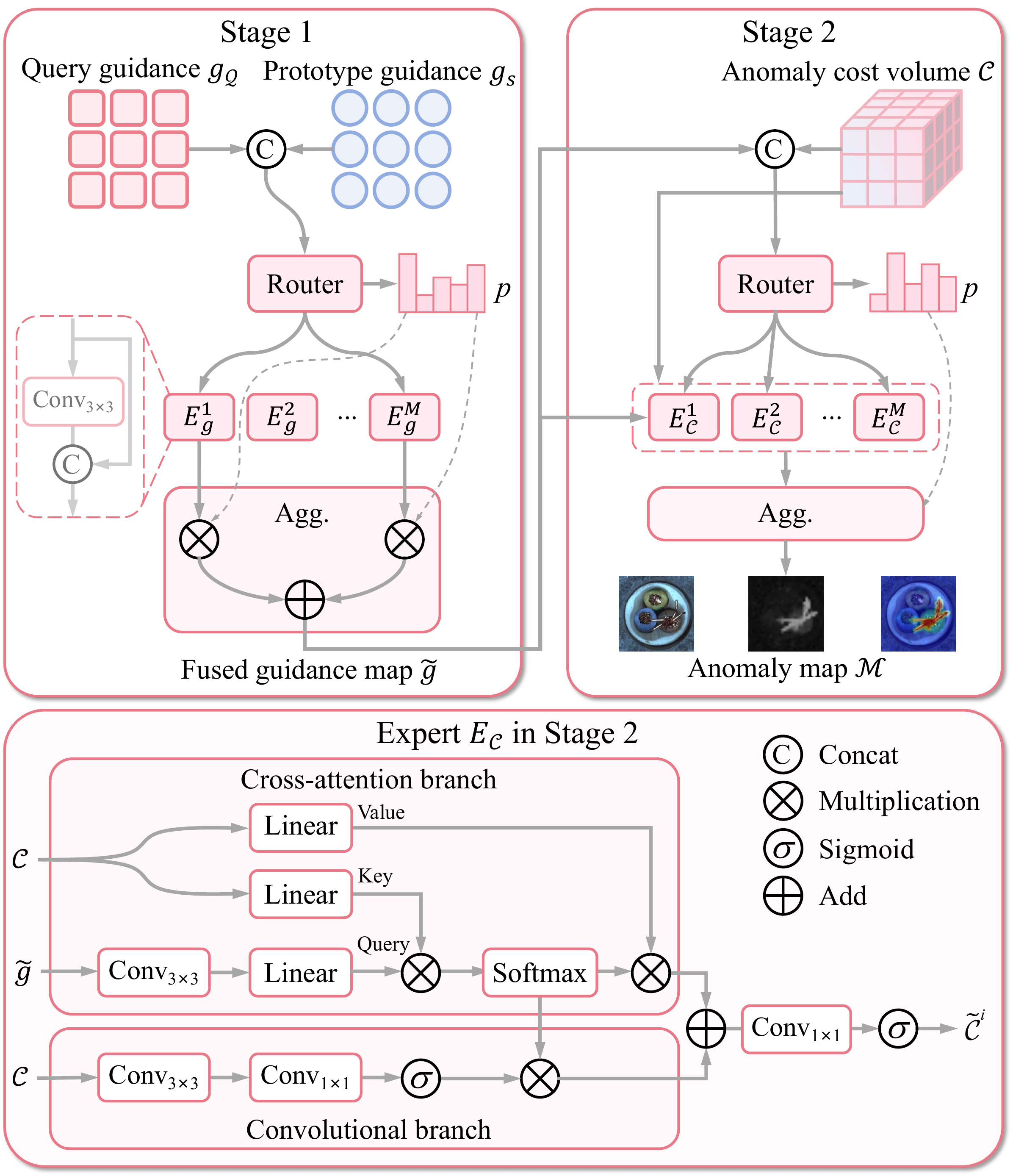}}
\vspace{-2mm}
\caption{Detailed architecture of the guided MoE filter.}
\label{fig:moearchitec}
\end{center}
\vspace{-8mm}
\end{figure}

The proposed guided MoE filter consists of two stages. The first stage generates a fused guidance map through dual-guidance aggregation, while the second stage filters the anomaly cost volume under this fused guidance. Fig.~\ref{fig:moearchitec} provides an overview of the detailed architecture.

In the first stage (Fig.~\ref{fig:moearchitec} (top left)), each expert $E_g^i$ is implemented as a lightweight residual block composed of a $3\times3$ convolution and a skip connection that concatenates the convolutional output with the expert input $\text{cat}(g_\mathcal{Q}, g_s)$. The gated ensemble of all guidance experts produces the fused guidance feature $\tilde{g}\in\mathbb{R}^{H'\times W'\times3D}$.

In the second stage (Fig.~\ref{fig:moearchitec} (top right)), the initial anomaly cost volume $\mathcal{C}$ is refined by modeling its semantic affinity with the fused guidance $\tilde{g}$. The concatenated feature $\text{cat}(\tilde{g}, \mathcal{C})$ is passed into a router that generates a dense soft gating distribution over all filtering experts:
$$
p = \text{Softmax}(\text{Router}(\text{cat}(\tilde{g}, \mathcal{C}))).
$$
Each denoising expert $E_\mathcal{C}^i$ (Fig.~\ref{fig:moearchitec} (bottom)) performs dual-branch filtering, i.e., a semantic-guided cross-attention branch and a confidence-aware convolution branch, to refine the initial anomaly cost volume $\mathcal{C}$ by modeling the semantic affinity with $\tilde{g}$.

\textbf{Cross-attention branch.} The fused guidance $\tilde{g}$ is convolved and projected into query space, while the cost volume $\mathcal{C}$ provides keys and values. The attention map $\mathcal{A}^i$ encodes semantic affinity and enables a semantically weighted refinement:
$$
\begin{aligned}
&\mathcal{A}^i=\text{Softmax}\!\left( \frac{(\text{Conv}_{3\times3}(\tilde{g})W^i_Q)(\mathcal{C}W^i_K)^\top}{\sqrt{K}} \right),\\
&\text{CrossAtt}(\tilde{g}, \mathcal{C}) = \mathcal{A}^i (\mathcal{C} W^i_V ),
\end{aligned}
$$
where $\text{Conv}_{3\times3}(\cdot)$ denotes a $3\times3$ convolutional layer, $W^i_Q,W^i_K,W^i_V\in\mathbb{R}^{K\times K}$ are learnable projections, $\mathcal{A}^i\in\mathbb{R}^{(H'W')\times (H'W')}$ is the attention map, and $\text{CrossAtt}(\tilde{g}, \mathcal{C})$ denotes the output of the cross-attention branch.

\begin{figure*}
\begin{center}
\centerline{\includegraphics[width=0.999\linewidth]{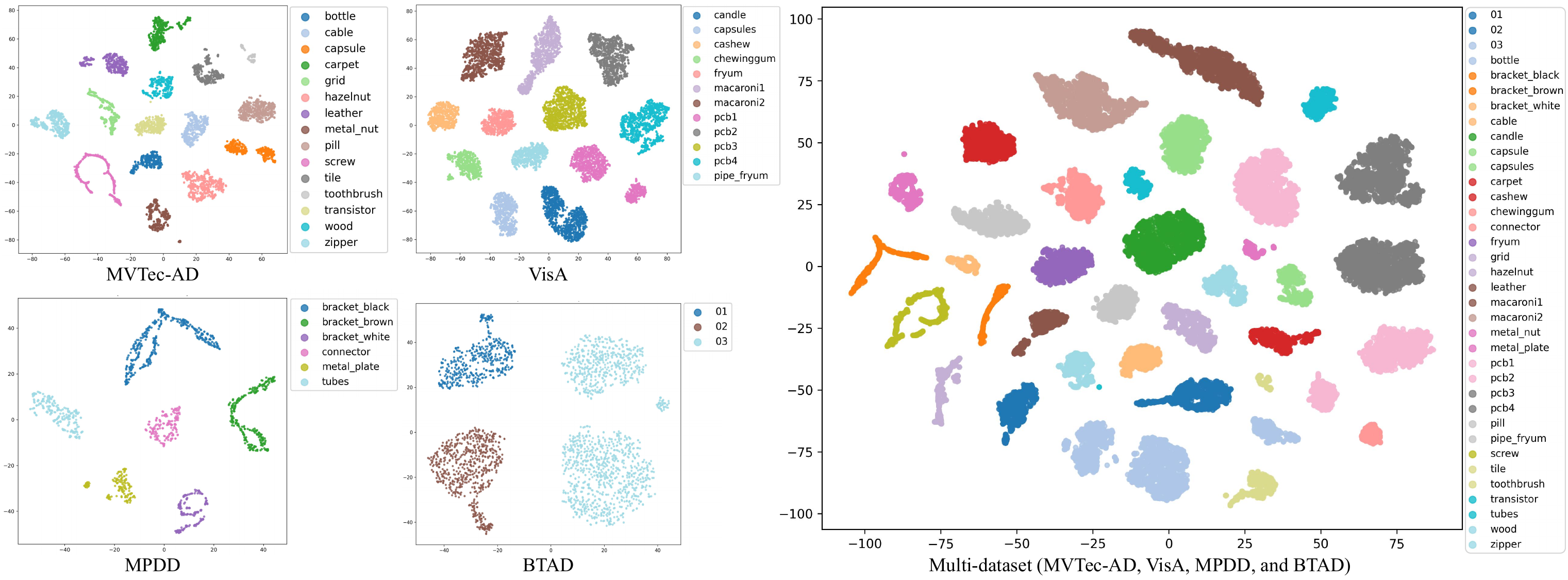}}
\vspace{-2mm}
\caption{T-SNE visualizations of CLS tokens on MVTec, VisA, MPDD, BTAD, and multi-dataset. Distinct and compact clusters reveal that CLS tokens encode strongly discriminative class-level semantics.}
\label{fig:tsne}
\end{center}
\vspace{-9mm}
\end{figure*}

\textbf{Convolutional branch.} This branch estimates and modulates the reliability of the refined matches. A confidence map $R^i\in\mathbb{R}^{H'\times W'\times K}$ is obtained from $\mathcal{C}$ through a pair of convolutional layers:
$$
R^i = \text{Sigmoid}(\text{Conv}_{1\times1}(\text{Conv}_{3\times3}(\mathcal{C}))).
$$
The confidence map is then used to modulate the attention response from the cross-attention branch via $\mathcal{A}^i R^i$.

Finally, each denoising expert $E_\mathcal{C}^i$ outputs a weighted combination of its two branches:
$$
\mathcal{\tilde{C}}^i = \text{Sigmoid}(\text{Conv}_{1\times1}(\text{CrossAtt}(\tilde{g}, \mathcal{C})+\beta\cdot (\mathcal{A}^iR^i))),
$$
where $\beta=0.1$ controls the influence of the reliability modulation.
The final anomaly map $\mathcal{M}$ is obtained by aggregating the expert outputs using the gating weights,
yielding a spatially coherent, semantically guided anomaly prediction.

\section{Feasibility of Category-Agnostic Retrieval}\label{supp:retrieval}
To assess whether templates from multiple datasets and categories can be jointly organized within our hierarchical vector database using class-level semantic representations, we visualize the CLS tokens (extracted from the DINOv2-s encoder~\cite{dinov2}) using t-distributed Stochastic Neighbor Embedding (t-SNE) on MVTec, VisA, MPDD, BTAD, and multi-dataset , incorporating all samples from each dataset (as shown in Fig.~\ref{fig:tsne}).

Across all datasets, the embeddings exhibit compact (intra-class) and well-separated clusters with clear inter-class boundaries, indicating that CLS tokens reliably encode global semantic cues that remain discriminative and stable across diverse categories. This also confirms that the derived class prototype entities are able to serve as robust semantic anchors that faithfully represent each category. These observations provide strong support for adopting CLS tokens as the foundational representation for class-level grouping within our hierarchical database, ensuring that each query can consistently locate its correct semantic cluster during category- and dataset-agnostic retrieval.

\begin{table}[t]
    \centering
    \tabcolsep=0.1cm
    \caption{Multi-class anomaly detection/localization performance of RAID integrated with reconstruction-based methods on MVTec-AD. All results are obtained by directly applying the full-shot \textbf{pretrained RAID model} without any additional fine-tuning. $^\dag$ denotes results using DINOv2-s with a $8\times8$ patch size.}
    \label{tab:tab_mvtec_Reconstructe}
    \vspace{-1mm}
    \footnotesize
    \begin{tabular}{l|c c cccc}
        \toprule
        Method & I-AUROC & I-AP& I-F1max & P-AUROC & P-AP & P-F1max \\
        \midrule
        GLAD$^\dag$~\cite{glad} & 97.5 & 98.8 & 96.8 &97.3&58.8 & 59.7\\
         \cellcolor{gray!20}+\ RAID$^\dag$ & \cellcolor{gray!20}\cellcolor{gray!20}\textbf{98.2}&\cellcolor{gray!20}\textbf{99.2}&\cellcolor{gray!20}\textbf{97.0} & \cellcolor{gray!20}\textbf{97.5} & \cellcolor{gray!20}\textbf{68.2} & \cellcolor{gray!20}\textbf{66.0}\\
        \midrule
         DiAD~\cite{diad} & 97.2 & \textbf{99.0} & \textbf{96.5} &96.8&52.6 & 55.5\\
          \cellcolor{gray!20}+\ RAID$^\dag$ & \cellcolor{gray!20}\cellcolor{gray!20}\textbf{97.4}&\cellcolor{gray!20}98.8&\cellcolor{gray!20}96.3 & \cellcolor{gray!20}\textbf{97.7} & \cellcolor{gray!20}\textbf{69.8} & \cellcolor{gray!20}\textbf{67.2}\\
        \bottomrule
    \end{tabular}
    \vspace{-3mm}
\end{table}

\section{Applicability to Reconstruction-based Approaches}\label{supp:recon}
In this section, we show that RAID can be seamlessly integrated with reconstruction-based UAD methods such as DiAD~\cite{diad} and GLAD~\cite{glad}. During inference, the reconstructed normal image of each test sample is treated as an exemplar for building a dynamic, sample-specific hierarchical vector database. In our experiments, RAID uses features from the final layer of DINO, whereas GLAD \cite{glad} and DiAD \cite{diad} rely on multi-layer features (four and two layers, respectively). Notably, we directly reuse the full-shot pretrained RAID weights without any fine-tuning. 

As reported in Table~\ref{tab:tab_mvtec_Reconstructe}, incorporating GLAD or DiAD into RAID yields substantial performance gains. These results demonstrate that RAID strengthens matching reliability and serves as a flexible, plug-and-play module for enhancing reconstruction-based anomaly detection methods.


\section{Ablation studies of Retrieval}
\label{sec:ablation_retrieval}
To systematically examine the impact of the retrieval process, we conduct targeted ablation studies on VisA focusing on two aspects: retrieval hierarchy and retrieval hyperparameters.

\textbf{Retrieval Hierarchy.} Removing class proto., semantic proto., and instance tokens as shown in Table \ref{tab:tab_retrieval_hierarchy} causes I-AUROC/P-AP drops of 1.6\%/2.0\%, 3.0\%/5.2\%, and 15.4\%/15.1\%, respectively, with inference times of 0.28s, 0.08s, and 0.07s per image.
Specifically, cases are sensitive to different levels: \emph{Capsules}, \emph{Macaroni1}, and \emph{PCB1} drop by 2.9\%/1.6\% (w/o class), 7.3\%/2.9\% (w/o semantic), and 41.9\%/67.4\% (w/o instance), respectively, highlighting the unique contributions of each level.

\begin{table}[t]
    \centering
    \footnotesize
    \caption{Effectiveness of retrieval hierarchy on VisA using I-AUROC/P-AP.}
    \label{tab:tab_retrieval_hierarchy}
    \vspace{-2mm}
    \begin{tabular}{c|c|c|c|c}
        \toprule
      Class & Semantic & Instance & Result(\%) & Inf.(s) \\
      \midrule
     -&\checkmark &\checkmark & 93.3 / 43.2 & 0.28\\ 
     \checkmark & - &\checkmark & 91.9 / 40.0 & 0.08\\
     \checkmark & \checkmark &- & 79.5 / 30.1 & 0.07\\
     \checkmark &  \checkmark & \checkmark & 94.9 / 45.2 & 0.05\\
     
        \bottomrule
    \end{tabular}
    \vspace{-3mm}
\end{table}

\textbf{Retrieval Hyperparameters.} 
We evaluate the sensitivity of retrieval hyperparameters, including clustering seeds, numbers of semantic prototypes (\#$\bar{\textbf{s}}$), retrieved semantic prototypes ($K’$) and retrieved patch tokens ($K$). Tables \ref{tab:tab_retrieval_Hyperparameters} and \ref{tab:tab_seeds} demonstrate our stable performance (robustness) across the first three settings. Performance drops only with insufficient patch retrieval. 

Class-level clustering is inherently stable due to large semantic separation between object categories, naturally grouping similar samples and making class prototypes insensitive to initialization. Semantic-level prototypes are derived from many patch tokens, providing sufficient coverage of intra-class variations. Under this regime, the semantic representations remain consistent across different configurations.

By contrast, retrieved patch tokens directly determine the informativeness of the cost volume, which drives MoE routing and guided filtering during generation. Retrieving too few tokens produces an under-informative cost volume, limiting expert routing and denoising, whereas retrieving sufficient tokens enables stable performance and eventual saturation. This highlights that retrieval is tightly coupled with the generation process.

\begin{table}[t]
    \centering
    \footnotesize
    \caption{Ablation of retrieval hyperparameters on VisA using I-AUROC/P-AP.}
    \label{tab:tab_retrieval_Hyperparameters}
    \vspace{-2mm}
    \begin{tabular}{c|c|c|c}
        \toprule
      \#$\bar{\textbf{s}}$  & $K'$ & $K$ & Result(\%) \\
      \midrule
      10 & 1 & 5 & 91.3 / 39.6\\ 
     100 & 1 & 5 & 91.4 / 39.5\\
     50 & 20 & 150 & 94.9 / 45.3\\
     50 &  5 & 150 & 94.9 / 45.2 \\
     
        \bottomrule
    \end{tabular}
    \vspace{-1mm}
\end{table}

\begin{table}[t]
    \centering
    \footnotesize
    \caption{Ablation of clustering seeds on VisA using I-AUROC/P-AP.}
    \label{tab:tab_seeds}
    \vspace{-2mm}
    \begin{tabular}{c|c|c|c}
        \toprule
      Seed & 0 & 2 & 42 \\ 
      \midrule
     Metrics & 95.0 / 45.9 & 94.9 / 45.2 & 94.9 / 45.9\\
        \bottomrule
    \end{tabular}
    \vspace{-1mm}
\end{table}


\begin{figure*}
\begin{center}
\centerline{\includegraphics[width=0.8\linewidth]{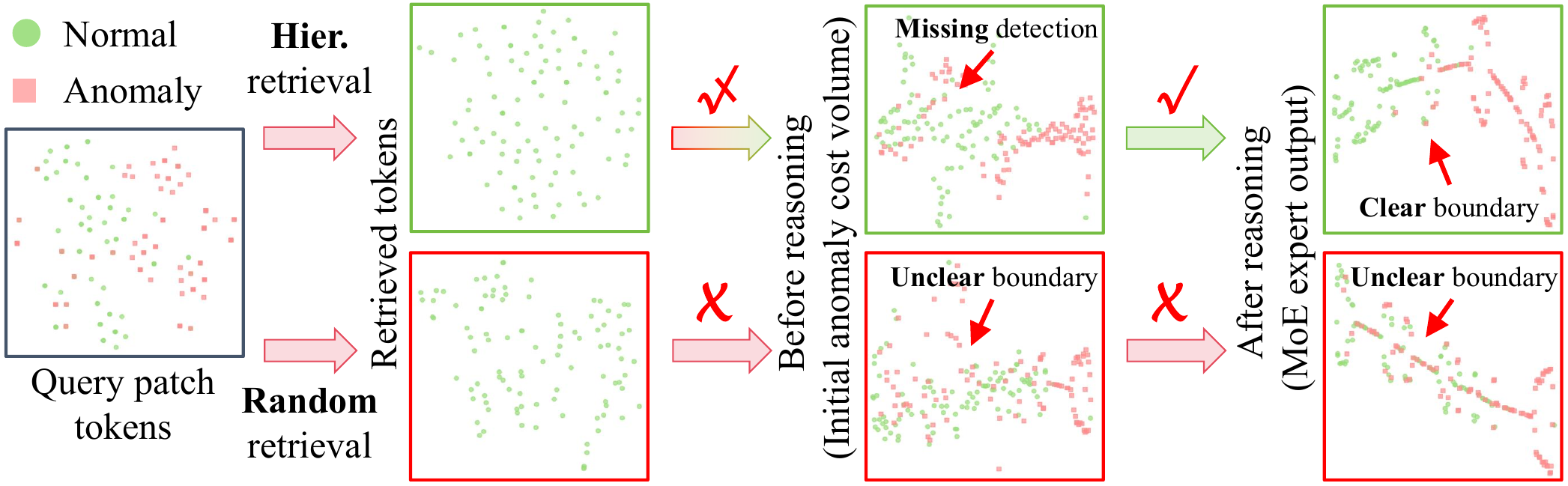}}
\vspace{-2mm}
\caption{T-SNE visualization of token-level representations before and after step-level reasoning.}
\label{fig:step_level}
\end{center}
\vspace{-2mm}
\end{figure*}

\section{Visualization of the Retrieval-reasoning Interaction}
\label{sec:retrieval-reasoning}
We provide t-SNE visualizations in Fig. \ref{fig:step_level} comparing token-level representations before \& after reasoning (step-level) under hierarchical vs. random retrieval, directly showing that \emph{retrieved evidence actively shapes intermediate representations rather than passively conditioning the model}.


\begin{figure}
\begin{center}
\centerline{\includegraphics[width=0.5\linewidth]{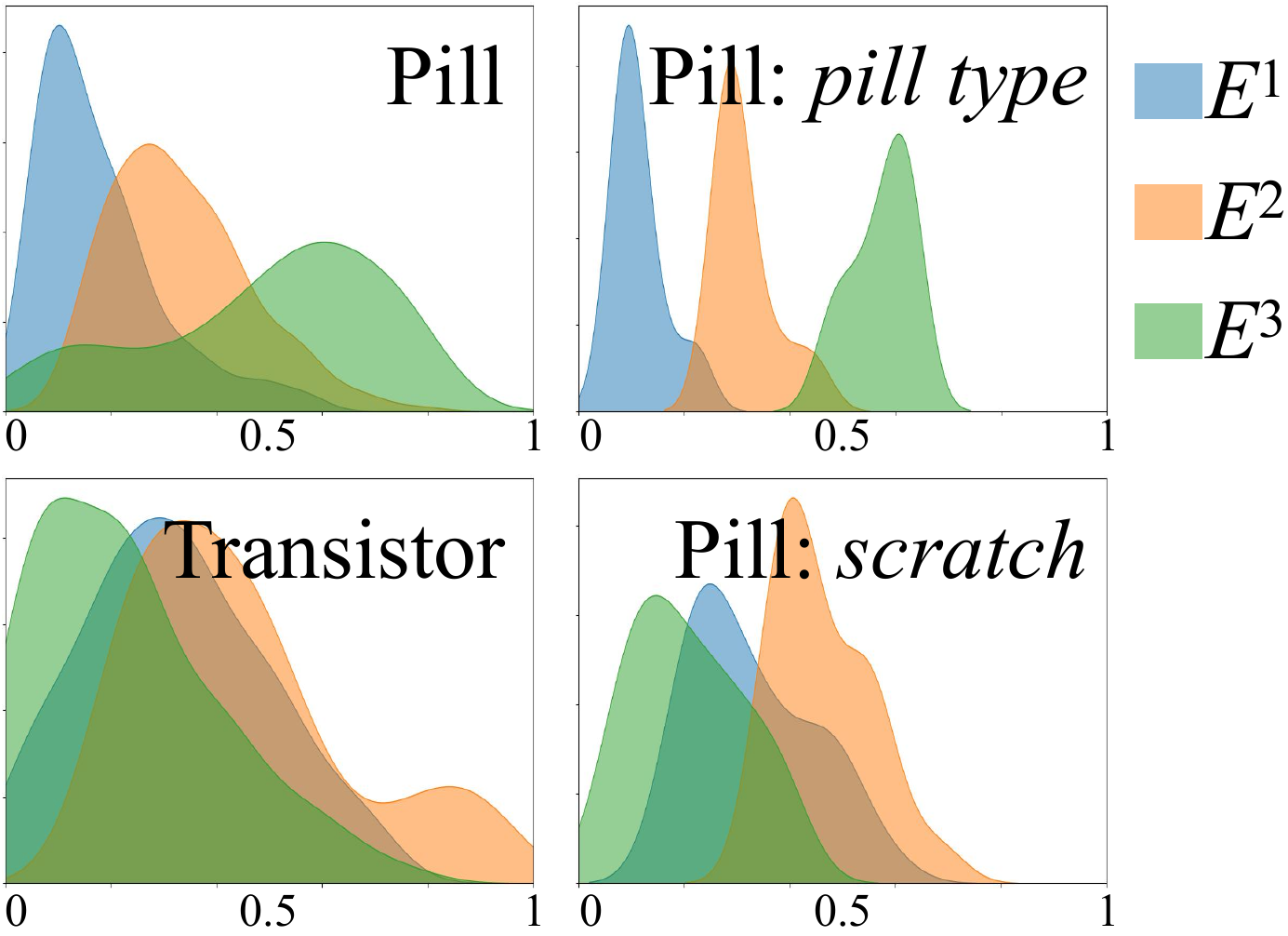}}
\vspace{-2mm}
\caption{KDE distributions of second-stage MoE routing weights across object categories and anomaly types.}
\label{fig:moe2}
\end{center}
\vspace{-8mm}
\end{figure}

\section{Visualizing Expert Specialization in MoE}
\label{sec:expert_specialization}
We analyze the routing behavior of the second-stage MoE. Specifically, we visualize the Kernel Density Estimation (KDE) distributions of routing weights across: different object categories (Pill and Transistor), and different anomaly types (pill type and scratch). The resulting distributions differ significantly across both dimensions in Fig. \ref{fig:moe2}, indicating that the router consistently assigns distinct importance profiles to experts depending on the semantic context and anomaly characteristics. This shows that, despite dense activation, experts contribute unequally and conditionally, forming a soft but structured specialization pattern.

\section{Ablation Studies on VisA}
\label{sec:ablation_study_visa}
We conduct ablations on the challenging VisA dataset, evaluating expert number, template quantity, and the guided MoE filter. Results in Tables \ref{tab:tab_expert_visa}, \ref{tab:tab_template_visa}, and \ref{tab:tab_moe_visa} consistently validate each component under realistic, non-saturated settings.

\begin{table}[t]
    \centering
    \footnotesize
    \caption{Effectiveness of expert quantity on VisA using I-AUROC/P-AP.}
    \label{tab:tab_expert_visa}
    \vspace{-2mm}
    \begin{tabular}{c|c|c|c}
        \toprule
      \#$(E_g, E_{\mathcal{C}})$ & (2, 3) & (3, 2) & (3, 3)\\
     \midrule
     Metrics & 94.0 / 40.5 & 94.1 / 41.8 & 94.9 / 45.2\\
        \bottomrule
    \end{tabular}
    \vspace{-1mm}
\end{table}

\begin{table}[t]
    \centering
    \footnotesize
    \tabcolsep=0.08cm
    \caption{Effectiveness of template quantity on VisA using I-AUROC/P-AP.}
    \label{tab:tab_template_visa}
    \vspace{-2mm}
    \begin{tabular}{c|c|c|c|c}
        \toprule
      \# template & 20 & 50 & 100 & All\\
     \midrule
     Metrics & 91.9 / 44.0 & 94.0 / 44.7 & 94.9 / 45.2 & 94.9 / 47.0\\
        \bottomrule
    \end{tabular}
    \vspace{-1mm}
\end{table}

\begin{table}[t]
    \centering
    \footnotesize
    \tabcolsep=0.02cm
    \caption{Ablation studies of guided MoE filter on VisA using I-AUROC/P-AP. IDs match with Table \ref{tab:tab_ablation_bi}.}
    \label{tab:tab_moe_visa}
    \vspace{-2mm}
    \begin{tabular}{c|c|c|c|c|c|c}
        \toprule
      ID & 1 & 2 & 3 & 4 & 5 & 7\\
     \midrule
     Metrics & 91.9 / 37.3 & 93.6 / 41.9 & 91.4 / 37.4 & 93.1 / 41.9 & 92.9 / 39.4 & 94.9 / 45.2\\
        \bottomrule
    \end{tabular}
    \vspace{-2mm}
\end{table}

\section{Detailed Quantitative Performance under the Full-shot Setting}\label{supp:results1}
In this section, we present detailed per-class results of our multi-class model for both image-level anomaly detection and pixel-level anomaly localization on the MVTec-AD and VisA datasets under the full-shot setting. Results for the 15 MVTec-AD categories are reported in Tables~\ref{tab:tab_mvtec_Auroc}, \ref{tab:tab_mvtec_Ap}, \ref{tab:tab_mvtec_I-f1max}, and \ref{tab:tab_mvtec_PRO}, while those for the 12 VisA categories are shown in Tables~\ref{tab:tab_visA_Auroc}, \ref{tab:tab_visA_AP}, \ref{tab:tab_visA_F1max}, and \ref{tab:tab_visA_PRO}.
Across both benchmarks, our approach consistently achieves strong performance in detection and localization metrics, demonstrating robustness and effectiveness across diverse anomaly categories and establishing state-of-the-art results.

\begin{table*}[t]
    \centering
\renewcommand{\arraystretch}{0.93}
    \tabcolsep=0.05cm
    \caption{Multi-class anomaly detection/localization performance on MVTec-AD with I-AUROC/P-AUROC metrics.}
    \label{tab:tab_mvtec_Auroc}
    \vspace{-2mm}
    \footnotesize
    \begin{tabular}{c| c |c| c |c |c| c|c|c |c|c|c}
        \toprule
        \multicolumn{2}{c|}{Category} & PatchCore  \cite{patchcore} & UniAD \cite{uniad}& SimpleNet & MambaAD \cite{mambaad} & GLAD \cite{glad}& DiAD \cite{diad} & ViTAD \cite{vitad} &   AnomalyDINO   &  Costfilter-AD & Ours \\
        \midrule
        
        \multirow{10}{*}{\rotatebox{90}{\shortstack{Object}}} 
        & Bottle & \textbf{100.0} / 99.2 & \textbf{99.7} / 98.1 & \textbf{100.0} / 97.2 & \textbf{100.0} / 98.8 & \textbf{100.0} / 99.7 & 98.4 / 98.4 & 99.5 / 99.0 & \textbf{100.0} / 99.3 & \textbf{100.0} / 98.8 & \textbf{100.0} / 99.3 \\
        & Cable & 95.3 / 93.6 & 95.2 / 97.3 & 97.5 / 96.7 & 98.8 / 95.8 & 98.7 / 93.4 & 94.8 / 96.8 & 99.7 / \textbf{98.6} & 99.6 / 98.3 & \textbf{99.8} / 98.2 & 98.9 / 97.5 \\
        & Capsule & 96.8 / 98.0 & 86.9 / 98.5 & 90.7 / 98.5 & 94.4 / 98.4 & 96.5 / 99.1 & 89.0 / 97.1 & \textbf{100.0} / \textbf{99.6} & 89.7 / 99.1 & 96.4 / 98.9 & 96.8 / 99.2 \\
        & Hazelnut & 99.3 / 97.6 & 99.8 / 98.1 & 99.9 / 98.4 & \textbf{100.0} / 99.0 & 97 / 98.9 & 99.5 / 98.3 & \textbf{100.0} / 96.6 & 99.9 / 99.6 & \textbf{100.0} / 99.2 & \textbf{100.0} / \textbf{99.7} \\
        & Metal\_nut & 99.1 / 96.3 & 99.2 / 62.7 & 96.9 / \textbf{98.0} & 99.9 / 96.7 & 99.9 / 97.3 & 99.1 / 97.3 & 98.7 / 96.4 & \textbf{100.0} / 96.7 & \textbf{100.0} / 97.9 & \textbf{100.0} / 96.5 \\
        & Pill & 86.4 / 90.8 & 93.7 / 95.0 & 88.2 / 96.5 & 97.0 / 97.4 & 94.4 / 97.9 & 95.7 / 95.7 & \textbf{100.0} / 98.8 & 97.2 / 98.1 & 96.9 / 96.5 & 99.2 / \textbf{99.4} \\
        & Screw & 94.2 / 98.9 & 87.5 / 98.3 & 76.7 / 96.5 & 94.7 / 99.5 & 93.4 / \textbf{99.6} & 90.7 / 97.9 & \textbf{98.5} / 96.2 & 74.3 / 97.6 & 95.3 / 99.0 & 97.9 / 99.4 \\
        & Toothbrush & \textbf{100.0} / 98.8 & 94.2 / 98.4 & 89.7 / 98.4 & 98.3 / 99.0 & 99.7 / 99.2 & 99.7 / 99.0 & 95.4 / 98.3 & 99.7 / 99.2 & \textbf{100.0} / 98.9 & \textbf{100.0} / \textbf{99.3 }\\
        & Transistor & 98.9 / 92.3 & 99.8 / 95.8 & 99.2 / 97.9 & \textbf{100.0} / 96.5 & 99.4 / 90.9 & 99.8 / 95.1 & 99.8 / \textbf{99.0} & 96.5 / 95.8 & \textbf{100.0} / 97.1 & 99.8 / 95.9 \\
        & Zipper & 97.1 / 95.7 & 95.8 / 96.8 & 99.0 / 97.9 & 99.3 / 98.4 & 96.4 / 93.0 & 95.1 / 96.2 & 99.7 / 96.4 & 98.8 / 94.3 & 98.9 / \textbf{98.3} & \textbf{99.7} / 98.1 \\
        \midrule
         \multirow{5}{*}{\rotatebox{90}{\shortstack{Texture}}} 
        & Carpet & 97.0 / 98.1 & 99.8 / 98.5 & 95.7 / 97.4 & 99.8 / 99.2 & 97.2 / 98.9 & 99.4 / 98.6 & 96.2 / 98.7 & 99.9 / 99.4 & \textbf{100.0} / 98.5 & \textbf{100.0} / \textbf{99.6 }\\
        & Grid & 91.4 / 98.4 & 98.2 / 63.1 & 97.6 / 96.8 & \textbf{100.0} / 99.2 & 95.1 / 98.2 & 98.5 / 96.6 & 91.3 / 99.0 & 98.7 / 97.8 & 99.3 / 98.3 & \textbf{100.0} / \textbf{99.4} \\
        & Leather & 100 / 99.2 & \textbf{100.0} / 98.8 & \textbf{100.0} / 98.7 & \textbf{100.0} / 99.4 & 99.5 / \textbf{99.7} & 99.8 / 98.8 & 98.9 / 99.1 & \textbf{100.0} / \textbf{99.7} & \textbf{100.0} / 99.3 & \textbf{100.0} / 99.4 \\
        & Tile & 96.0 / 90.3 & 99.3 / 91.8 & 99.3 / 95.7 & 98.2 / 93.8 & 100 / 97.8 & 96.8 / 92.4 & 98.8 / 93.9 & \textbf{100.0} / 98.5 & \textbf{100.0} / 95.0 & \textbf{100.0} / \textbf{99.3} \\
        & Wood & 93.8 / 90.8 & 98.6 / 93.2 & 98.4 / 91.4 & 98.8 / 94.4 & 95.4 / 96.8 & \textbf{99.7} / 93.3 & 97.6 / 95.9 & 97.9 / \textbf{97.6} & 98.5 / 94.3 & 98.4 / 97.4 \\
        \midrule
        \multicolumn{2}{c|}{Mean} & 96.4 / 95.7 & 96.5 / 96.8 & 95.3 / 96.9 & 98.6 / 97.7 & 97.5 / 97.3 & 97.2 / 96.8 & 98.3 / 97.7 & 96.8 / 98.1 & 99.0 / 98.0 & \textbf{99.4} / \textbf{98.6} \\
        \bottomrule
    \end{tabular}
    \vspace{-1mm}
\end{table*}

\begin{table*}[t]
    \centering
\renewcommand{\arraystretch}{0.92}
    \tabcolsep=0.11cm
    \caption{Multi-class anomaly detection/localization performance on MVTec-AD with I-AP/P-AP metrics.}
    \label{tab:tab_mvtec_Ap}
    \vspace{-2mm}
    \footnotesize
    \begin{tabular}{c |c |c |c |c| c|c|c |c|c|c}
        \toprule
        \multicolumn{2}{c|}{Category} & UniAD \cite{uniad} & SimpleNet \cite{simplenet} & MambaAD \cite{mambaad}& GLAD \cite{glad} & DiAD \cite{diad}& ViTAD \cite{vitad}& AnomalyDINO & Costfilter-AD & Ours \\
        \midrule
        
        \multirow{10}{*}{\rotatebox{90}{\shortstack{Object}}} 
        & Bottle & \textbf{100.0} / 60.0 & \textbf{100.0} / 53.8 & \textbf{100.0} / 79.7 & \textbf{100.0} / 80.9 & 96.5 / 52.2 & 99.9 / 60.5 & \textbf{100.0} / 87.3 & \textbf{100.0} / 81.4 & \textbf{100.0} / \textbf{89.6} \\
        & Cable & 95.9 / 39.9 & 98.5 / 42.4 & 99.2 / 42.2 & 99.3 / 51.4 & 98.8 / 50.1 & \textbf{99.9} / 31.2 & 99.8 / \textbf{69.3} & 99.8 / 58.0 & 99.4 / 62.7 \\
        & Capsule & 97.8 / 42.7 & 97.9 / 35.4 & 98.7 / 43.9 & 99.2 / 49.1 & 97.5 / 42.0 & \textbf{100.0} / 52.1 & 97.3 / 45.9 & 99.2 / 49.2 & 99.0 / \textbf{55.3} \\
        & Hazelnut & \textbf{100.0} / 55.2 & 99.9 / 44.6 & \textbf{100.0} / 63.6 & 98.2 / 68.0 & 99.7 / 79.2 & \textbf{100.0} / 56.4 & 99.9 / 79.0 & \textbf{100.0} / 72.4 & \textbf{100.0} / \textbf{87.5} \\
        & Metal\_nut & 99.9 / 14.6 & 99.3 / 83.1 & \textbf{100.0} / 74.5 & \textbf{100.0} / 81.8 & 96.0 / 30.0 & 99.6 / 60.6 & \textbf{100.0} / 77.2 & \textbf{100.0} / \textbf{79.0} & \textbf{100.0} / 74.9 \\
        & Pill & 98.7 / 44.0 & 97.7 / 72.4 & 99.5 / 64.0 & 99.0 / 73.9 & 98.5 / 46.0 & \textbf{100.0} / 79.9 & 99.5 / 78.6 & 99.4 / 59.6 & 99.9 / \textbf{84.2 }\\
        & Screw & 96.5 / 28.7 & 90.6 / 15.9 & 97.9 / 49.8 & 98.0 / 47.8 & \textbf{99.7} / \textbf{60.6} & 99.1 / 43.1 & 88.0 / 12.5 & 98.3 / 33.5 & 99.3 / 56.3 \\
        & Toothbrush & 97.4 / 34.9 & 95.7 / 46.9 & 99.3 / 48.5 & 99.9 / 45.0 & 99.9 / \textbf{78.7} & 99.0 / 42.7 & 99.9 / 46.9 & \textbf{100.0} / 51.6 & 99.9 / 71.1 \\
        & Transistor & 98.0 / 59.5 & 98.7 / 58.2 & \textbf{100.0} / 69.4 & 99.2 / 58.9 & 99.6 / 15.6 & 99.9 / 64.6 & 96.1 / 62.4 & \textbf{100.0} / \textbf{76.6} & 99.7 / 64.5 \\
        & Zipper & 99.5 / 40.1 & 99.7 / 53.4 & 99.8 / 60.4 & 98.9 / 40.9 & 99.1 / 60.7 & 99.9 / \textbf{75.1} & 99.7 / 44.0 & 99.7 / 55.1 & \textbf{99.9} / 59.9 \\
        \midrule
         \multirow{5}{*}{\rotatebox{90}{\shortstack{Texture}}} 
        & Carpet & 99.9 / 49.9 & 98.7 / 38.7 & 99.9 / 60.0 & 99.1 / 72.2 & 99.9 / 42.2 & 99.3 / 77.8 & \textbf{100.0} / 76.2 & \textbf{100.0} / 64.7 & \textbf{100.0} / \textbf{82.1} \\
        & Grid & 99.5 / 10.7 & 99.2 / 20.5 & \textbf{100.0} / 47.4 & 93.6 / 10.2 & 99.8 / \textbf{66.0 }& 97.0 / 34.0 & 99.3 / 31.0 & 99.8 / 34.0 & \textbf{100.0} / 52.1 \\
        & Leather & \textbf{100.0} / 32.9 & \textbf{100.0} / 28.5 & \textbf{100.0} / 50.3 & 99.8 / 61.7 & 99.7 / 56.1 & 99.6 / 51.3 & \textbf{100.0} / 60.2 & \textbf{100.0} / 47.4 & \textbf{100.0} / \textbf{62.2} \\
        & Tile & 99.8 / 42.1 & 99.8 / 60.5 & 99.3 / 45.1 & \textbf{100.0} / 70.3 & 99.9 / 65.7 & 98.3 / 58.4 & \textbf{100.0} / 76.4 & \textbf{100.0} / 56.0 & \textbf{100.0} / \textbf{93.7} \\
        & Wood & 99.6 / 37.2 & 99.5 / 34.8 & 99.6 / 46.2 & 98.5 / 70.6 & \textbf{100.0} / 43.3 & 99.3 / 42.6 & 99.3 / 72.7 & 99.5 / 52.4 & 99.5 / \textbf{80.4} \\
        \midrule
        \multicolumn{2}{c|}{Mean} & 98.8 / 43.4 & 98.4 / 45.9 & 99.6 / 56.3 & 98.8 / 58.8 & 99.0 / 52.6 & 99.4 / 55.3 & 98.6 / 61.3 & 99.7 / 58.1 & \textbf{99.8} / \textbf{71.7} \\
        \bottomrule
    \end{tabular}
    \vspace{-2mm}
\end{table*}

\section{Detailed Quantitative Performance under the Few-shot Setting}
In this section, we report the mean and standard deviation of per-category performance under few-shot settings, with all results averaged over five random seeds. During retrieval, template samples are augmented with rotations to enhance diversity. Tables~\ref{tab:few-shot-mvtec} and~\ref{tab:few-shot_visa} summarize the image-level and pixel-level AUROC (I-AUROC and P-AUROC) for each category. The stable improvements observed across all categories and both datasets further highlight the strong generalizability of our method in few-shot anomaly detection.

\begin{table*}[t]
    \centering
    \tabcolsep=0.11cm
    \caption{Multi-class anomaly detection/localization performance on MVTec-AD with I-F1max/P-F1max metrics.}
    \label{tab:tab_mvtec_I-f1max}
    \footnotesize
    \begin{tabular}{c |c |c |c |c| c|c |c|c|c|c}
        \toprule
        \multicolumn{2}{c|}{Category} & UniAD \cite{uniad} & SimpleNet \cite{simplenet}& MambaAD \cite{mambaad}& GLAD \cite{glad}& DiAD \cite{diad}& ViTAD \cite{vitad}& AnomalyDINO & Costfilter-AD & Ours \\
        \midrule
        
        \multirow{10}{*}{\rotatebox{90}{\shortstack{Object}}} 
        & Bottle & \textbf{100.0} / 69.2 & \textbf{100.0} / 62.4 & \textbf{100.0} / 76.7 & \textbf{100.0} / 75.5 & 91.8 / 54.8 & 99.4 / 64.1 & \textbf{100.0} / 80.2 & \textbf{100.0} / 77.6 & \textbf{100.0} / \textbf{83.0} \\
        & Cable & 88.0 / 45.2 & 94.7 / 51.2 & 95.7 / 48.1 & 97.3 / 53.4 & 95.2 / 57.8 & \textbf{99.1} / 36.7 & 97.8 / \textbf{67.0} & 98.4 / 63.0 & 96.1 / 62.0 \\
        & Capsule & 94.4 / 46.5 & 93.5 / 44.3 & 94.9 / 47.7 & 96.8 / 51.2 & 95.5 / 45.3 & \textbf{100.0} / 55.8 & 94.3 / 48.9 & 96.4 / 53.0 & 98.6 / \textbf{57.4} \\
        & Hazelnut & 99.3 / 56.8 & 99.1 / 51.4 & \textbf{100.0} / 64.4 & 94.4 / 63.8 & 97.3 / 80.4 & \textbf{100.0} / 68.8 & 99.3 / 75.5 & \textbf{100.0} / 70.8 & 99.3 / \textbf{82.8} \\
        & Metal\_nut & 99.5 / 29.2 & 96.1 / 79.4 & 99.5 / 79.1 & 99.5 / \textbf{82.4} & 91.6 / 38.3 & 96.7 / 58.3 & \textbf{100.0} / 79.5 & 99.5 / 82.1 & \textbf{100.0} / 74.2 \\
        & Pill & 95.7 / 95.3 & 53.9 / 67.7 & 96.2 / 66.5 & 94.6 / 69.9 & 94.5 / 51.4 & \textbf{100.0} / 75.6 & 97.1 / 71.1 & 96.9 / 61.2 & 98.6 / \textbf{78.7} \\
        & Screw & 89.0 / 92.6 & 37.6 / 23.2 & 94.0 / 50.9 & 92.2 / 47.6 & \textbf{97.9} / \textbf{59.6} & 95.7 / 47.4 & 87.2 / 19.4 & 94.5 / 40.5 & 96.3 / 54.1 \\
        & Toothbrush & 95.2 / 45.7 & 92.3 / 52.5 & 98.4 / 59.2 & 98.4 / 57.4 & 99.2 / \textbf{72.8} & 95.5 / 47.8 & 98.4 / 57.7 & \textbf{100.0} / 61.9 & 98.4 / 71.2 \\
        & Transistor & 93.8 / 64.6 & 97.6 / 56.0 & \textbf{100.0} / 67.1 & 95.0 / 58.3 & 97.4 / 31.7 & 98.6 / 64.0 & 89.7 / 59.5 & \textbf{100.0} / \textbf{74.1} & 97.6 / 59.5 \\
        & Zipper & 97.1 / 49.9 & 98.3 / 54.6 & 97.5 / 61.7 & 95.6 / 46.2 & 94.4 / 60.0 & 98.4 / \textbf{77.3} & 97.9 / 49.3 & 98.3 / 59.5 & \textbf{99.2} / 59.5 \\
        \midrule
         \multirow{5}{*}{\rotatebox{90}{\shortstack{Texture}}} 
        & Carpet & 99.4 / 51.1 & 93.2 / 43.2 & 99.4 / 63.3 & 96.6 / 67.9 & 98.3 / 46.4 & 96.4 / 75.2 & 99.4 / 67.7 & \textbf{100.0} / 63.3 & 99.4 / \textbf{75.8} \\
        & Grid & 97.3 / 11.9 & 96.4 / 27.6 & \textbf{100.0} / 47.7 & 98.3 / 24.1 & 97.7 / \textbf{64.1} & 93.0 / 41.0 & 96.6 / 37.4 & 98.2 / 40.6 & \textbf{100.0} / 54.1 \\
        & Leather & \textbf{100.0} / 34.4 & \textbf{100.0} / 32.9 & \textbf{100.0} / 53.3 & 98.4 / 60.7 & 97.6 / \textbf{62.3} & 96.8 / 61.9 & \textbf{100.0} / 57.4 & \textbf{100.0} / 50.0 & 99.5 / 58.3 \\
        & Tile & 98.2 / 50.6 & 98.8 / 59.9 & 95.4 / 54.8 & \textbf{100.0} / 71.5 & 98.4 / 64.1 & 92.5 / 55.3 & \textbf{100.0} / 76.6 & \textbf{100.0} / 63.5 & \textbf{100.0} / \textbf{85.3} \\
        & Wood & 96.6 / 41.5 & 96.7 / 39.7 & 96.6 / 48.2 & 95.1 / 65.2 & \textbf{100.0} / 43.5 & 97.1 / 50.8 & 98.4 / 65.4 & 97.5 / 56.5 & 97.6 / \textbf{72.0} \\
        \midrule
        \multicolumn{2}{c|}{Mean} & 96.4 / 49.5 & 95.8 / 49.7 & 97.6 / 59.2 & 96.8 / 59.7 & 96.5 / 55.5 & 97.3 / 58.7 & 97.1 / 60.8 & 98.6 / 61.2 & \textbf{98.7} / \textbf{68.5} \\
        \bottomrule
    \end{tabular}
    \vspace{-1mm}
\end{table*}

\begin{table*}[t]
    \centering
\renewcommand{\arraystretch}{0.92}
    \tabcolsep=0.09cm
    \caption{Multi-class anomaly localization performance on MVTec-AD with AUPRO metrics.}
    \label{tab:tab_mvtec_PRO}
    \footnotesize
    \begin{tabular}{c |c |c |c |c| c|c|c |c|c|c}
        \toprule
        \multicolumn{2}{c|}{Category} & UniAD \cite{uniad} & SimpleNet \cite{simplenet}& MambaAD \cite{mambaad}& GLAD \cite{glad}& DiAD \cite{diad}& ViTAD \cite{vitad}& AnomalyDINO \cite{Anomalydino} & Costfilter-AD \cite{CostFilterAD}& Ours \\
        \midrule
        
        \multirow{10}{*}{\rotatebox{90}{\shortstack{Object}}} 
        & Bottle & 93.1 & 89.0 & 95.2 & 96.1 & 86.6 & 94.7 & 97.5 & 96.1 & \textbf{97.1} \\
        & Cable & 86.1 & 85.4 & 90.3 & 89.6 & 80.5 & \textbf{95.8} & 94.2 & 91.7 & 91.4 \\
        & Capsule & 92.1 & 84.5 & 92.6 & 96.1 & 87.2 & \textbf{97.9} & 95.8 & 92.9 & 96.4 \\
        & Hazelnut & 94.1 & 87.4 & 95.7 & 90.8 & 91.5 & 87.0 & 92.5 & 92.9 & \textbf{97.6} \\
        & Metal\_nut & 81.8 & 85.2 & 93.7 & 94.2 & 90.6 & 88.0 & 94.7 & 93.3 & \textbf{95.8} \\
        & Pill & 95.3 & 81.9 & 95.7 & 94.3 & 89.0 & 94.3 & 96.7 & 96.1 & \textbf{98.2} \\
        & Screw & 87.9 & 84.0 & \textbf{97.1} & 96.7 & 95.0 & 90.2 & 89.4 & 95.4 & 96.2 \\
        & Toothbrush & 93.5 & 87.4 & 91.7 & 95.6 & 95.0 & 92.0 & 96.1 & 89.6 & \textbf{96.8} \\
        & Transistor & \textbf{97.9} & 83.2 & 87.0 & 86.5 & 90.0 & 95.2 & 84.2 & 93.7 & 82.8 \\
        & Zipper & 92.6 & 90.7 & \textbf{94.3} & 84.5 & 91.6 & 92.4 & 86.2 & 93.3 & 93.5 \\
        \midrule
         \multirow{5}{*}{\rotatebox{90}{\shortstack{Texture}}} 
        & Carpet & 94.4 & 90.6 & 96.7 & 95.3 & 90.6 & 95.3 & 97.6 & 95.4 & \textbf{98.3} \\
        & Grid & 92.9 & 97.0 & 96.4 & 92.7 & 94.0 & 93.5 & 90.0 & 93.4 & \textbf{97.3} \\
        & Leather & 96.8 & 98.7 & 93.0 & 97.0 & 91.3 & 90.9 & 98.5 & \textbf{98.8} & 97.3 \\
        & Tile & 78.4 & 80.0 & 93.0 & 96.8 & 90.7 & 76.8 & 96.7 & 85.5 & \textbf{97.3} \\
        & Wood & 86.7 & 91.2 & 95.9 & 86.3 & 97.5 & 87.2 & 93.4 & 90.0 & \textbf{96.1} \\
        \midrule
        \multicolumn{2}{c|}{Mean} & 90.7 & 93.1 & 93.1 & 92.8 & 90.7 & 91.4 & 93.6 & 93.2 & \textbf{95.5} \\
        \bottomrule
    \end{tabular}
    \vspace{-1mm}
\end{table*}

\begin{table*}[t]
    \centering
    \tabcolsep=0.09cm
    \footnotesize
    \caption{Multi-class anomaly detection/localization on VisA with I-AUROC/P-AUROC metrics.}
    \label{tab:tab_visA_Auroc}
    \begin{tabular}{c|c|c|c|c|c|c|c|c|c|c}
        \toprule
        \multicolumn{2}{c|}{Category} & UniAD \cite{uniad}& SimpleNet \cite{simplenet} & MambaAD \cite{mambaad}& DiAD \cite{diad} & GLAD \cite{glad} & ViTAD \cite{vitad}& AnomalyDINO & Costfilter-AD & Ours \\
        \midrule
       \multirow{4}{*}{\rotatebox{90}{\shortstack{Complex\\Structure}}} 
            & PCB1 & 92.8 / 93.3 & 91.6 / 99.2 & 95.4 / \textbf{99.8} & 88.1 / 98.7 & 69.9 / 97.6 & 95.8 / 99.5 & 87.4 / 99.3 & \textbf{96.3} / 99.3 & 92.0 / 99.4 \\
            & PCB2 & 87.8 / 93.9 & 92.4 / 96.6 & 94.2 / \textbf{98.9} & 91.4 / 95.2 & 89.9 / 97.1 & 90.6 / 97.9 & 81.9 / 94.2 & \textbf{97.0} / 98.0 & 92.3 / 98.1 \\
            & PCB3 & 78.6 / 97.3 & 89.1 / 97.2 & 93.7 / \textbf{99.1} & 86.2 / 96.7 & 93.3 / 96.2 & 90.9 / 98.2 & 87.4 / 96.5 & 89.8 / 97.7 & \textbf{95.3} / 97.9 \\
            & PCB4 & 98.8 / 94.9 & 97.0 / 93.9 & \textbf{99.9} / 98.6 & 99.6 / 97.0 & 99.0 / \textbf{99.4} & 99.1 / 99.1 & 96.7 / 97.3 & 98.7 / 97.8 & 96.1 / 98.1 \\
        \midrule
        \multirow{4}{*}{\rotatebox{90}{\shortstack{Multiple\\instances}}}
            & Macaroni 1 & 79.9 / 97.4 & 85.9 / 98.9 & 91.6 / 99.5 & 85.7 / 94.1 & 93.1 / \textbf{99.9} & 85.8 / 98.5 & 88.0 / 98.2 & \textbf{93.7} / 99.4 & 92.8 / 99.6 \\
            & Macaroni 2 & 71.6 / 95.2 & 68.3 / 93.2 & 81.6 / 99.5 & 62.5 / 93.6 & 74.5 / 99.5 & 79.1 / 98.1 & 75.9 / 96.9 & 88.3 / 98.5 & \textbf{90.6} / \textbf{99.6} \\
            & Capsules & 55.6 / 88.7 & 74.1 / 97.1 & 91.8 / 99.1 & 58.2 / 97.3 & 88.8 / \textbf{99.3} & 79.2 / 98.2 & 93.6 / 97.0 & 80.1 / 97.6 & \textbf{96.6} / 99.1 \\
            & Candle & 94.1 / 98.5 & 84.1 / 97.6 & 96.8 / 99.0 & 92.8 / 97.3 & 86.4 / 98.8 & 90.4 / 96.2 & 90.3 / 96.1 & \textbf{97.8} / 99.2 & 96.0 / \textbf{99.6} \\
        \midrule
        \multirow{4}{*}{\rotatebox{90}{\shortstack{Single\\instances}}}
           & Cashew & 92.8 / 98.6 & 88.0 / 98.9 & 94.5 / 94.3 & 91.5 / 90.9 & 92.6 / 86.2 & 87.8 / 98.5 & \textbf{95.1} / 99.2 & 94.1 / 99.3 & 94.8 / \textbf{99.4} \\
            & \scriptsize{Chewing gum} & 96.3 / 98.8 & 96.4 / 97.9 & 97.7 / 98.1 & 99.1 / 94.7 & 98.0 / \textbf{99.6 }& 94.9 / 97.8 & 98.0 / 99.3 & \textbf{99.3} / 99.5 & 98.9 / \textbf{99.6} \\
            & Fryum & 83.0 / 95.9 & 88.4 / 93.0 & 95.2 / 96.9 & 89.8 / 97.6 & \textbf{97.2} / 96.8 & 94.3 / 97.5 & 93.4 / 96.1 & 88.9 / 97.8 & 94.6 / \textbf{98.1} \\
            & Pipe fryum & 94.7 / 98.9 & 90.8 / 98.5 & 98.7 / 99.1 & 96.2 / 99.4 & 98.1 / 98.9 & 97.8 / \textbf{99.5} & 98.0 / 99.1 & 96.6 / \textbf{99.5} & \textbf{98.8} / 99.1 \\
        \midrule
        \multicolumn{2}{c|}{Mean} & 85.5 / 95.9 & 87.2 / 96.8 & 94.3 / 98.5 & 86.8 / 96.0 & 90.1 / 97.4 & 90.5 / 98.2 & 90.5 / 97.5 & 93.4 / 98.6 & \textbf{94.9} / \textbf{99.0} \\
        \bottomrule
    \end{tabular}
    \vspace{-1mm}
\end{table*}

\begin{table*}[t]
    \centering
    \tabcolsep=0.09cm
    \footnotesize
    \caption{Multi-class anomaly detection/localization on VisA with I-AP/P-AP metrics.}
    \label{tab:tab_visA_AP}
    \begin{tabular}{c|c|c|c|c|c|c|c|c|c|c}
        \toprule
        \multicolumn{2}{c|}{Category} & UniAD \cite{uniad}& SimpleNet \cite{simplenet} & MambaAD \cite{mambaad}& DiAD \cite{diad} & GLAD \cite{glad} & ViTAD \cite{vitad}& AnomalyDINO & Costfilter-AD & Ours \\
        \midrule
       \multirow{4}{*}{\rotatebox{90}{\shortstack{Complex\\Structure}}} 
            & PCB1 & 92.7 / 83.6 & 91.9 / \textbf{86.1} & \textbf{93.0} / 77.1 & 88.7 / 49.6 & 72.5 / 38.0 & 94.7 / 64.5 & 84.6 / 81.3 & 91.6 / 66.9 & 91.1/78.9 \\
            & PCB2 & 87.7 / 4.2 & 93.3 / 8.9 & \textbf{93.7} / 13.3 & 91.4 / 7.5 & 88.9 / 6.4 & 89.9 / 12.6 & 81.1 / 12.0 & 92.0 / \textbf{21.1} & 93.4 / \textbf{21.1} \\
            & PCB3 & 78.6 / 13.8 & 91.1 / \textbf{31.0} & 94.1 / 18.3 & 87.6 / 8.0 & 94.0 / 25.0 & 91.2 / 22.4 & 90.2 / 23.3 & 82.1 / 28.6 & \textbf{95.4} / 26.2 \\
            & PCB4 & 98.8 / 14.7 & 97.0 / 23.9 & \textbf{99.9} / 47.0 & 99.5 / 17.6 & 98.2 / \textbf{52.6} & 98.9 / 42.9 & 96.3 / 37.4 & 97.1 / 39.4 & 94.6 / 40.3 \\
        \midrule
        \multirow{4}{*}{\rotatebox{90}{\shortstack{Multiple\\instances}}}
            & Macaroni 1 & 79.8 / 3.7 & 82.5 / 3.5 & 89.8 / 17.5 & 85.2 / 10.2 & \textbf{93.1} / 11.0 & 83.9 / 8.0 & 88.9 / 10.6 & 86.7 / \textbf{19.7} & 91.7 / 14.2 \\
            & Macaroni 2 & 71.6 / 0.9 & 54.3 / 0.6 & 78.0 / 9.2 & 57.4 / 0.9 & 73.8 / 7.0 & 74.7 / 3.6 & 76.2 / 5.5 & 81.5 / \textbf{15.5} & \textbf{91.7} / 13.1 \\
            & Capsules & 55.6 / 3.0 & 82.8 / 52.9 & 95.0 / \textbf{61.3} & 69.0 / 10.0 & 94.1 / 47.8 & 87.6 / 30.4 & \textbf{96.4} / 43.3 & 79.4 / 51.4 & 73.3 / 55.1 \\
            & Candle & 94.0 / 17.6 & 73.3 / 8.4 & 96.9 / 23.2 & 92.0 / 12.8 & 88.2 / 29.3 & 91.2 / 16.8 & 90.2 / 28.1 & 92.5 / 41.0 & \textbf{96.4} / \textbf{36.5} \\
        \midrule
        \multirow{4}{*}{\rotatebox{90}{\shortstack{Single\\instances}}}
           & Cashew & 92.8 / 51.7 & 91.3 / 68.9 & 97.3 / 46.8 & 95.7 / 53.1 & 96.4 / 29.2 & 94.2 / 63.9 & \textbf{97.6} / 60.2 & 91.5 / 66.1 & 97.4 / \textbf{70.3} \\
            & \scriptsize{Chewing gum} & 96.2 / 54.9 & 98.2 / 26.8 & 98.9 / 57.5 & \textbf{99.5} / 11.9 & 99.1 / 73.9 & 97.7 / 61.6 & 96.9 / 64.7 & \textbf{99.5} / 40.9 & \textbf{99.5} / \textbf{78.5} \\
            & Fryum & 83.0 / 34.0 & 93.0 / 39.1 & \textbf{97.7} / 47.8 & 95.0 / \textbf{58.6} & 98.9 / 36.1 & 97.4 / 47.1 & 97.4 / 46.7 & 86.8 / 54.0 & 97.6 / 52.4 \\
            & Pipe fryum & 94.7 / 50.2 & 95.5 / 65.6 & \textbf{99.3} / 53.5 & 98.1 / \textbf{72.7} & \textbf{99.3} / 50.1 & 99.0 / 66.0 & 99.2 / 61.0 & 93.5 / 71.3 & 99.3 / 56.2 \\
        \midrule
        \multicolumn{2}{c|}{Mean} & 85.5 / 18.4 & 87.0 / 31.5 & 94.5 / 37.6 & 88.3 / 20.3 & 91.4 / 33.9 & 91.7 / 36.6 & 91.4 / 39.6 & 89.3 / 45.0 & \textbf{95.5} / \textbf{45.2} \\
        \bottomrule
    \end{tabular}
    \vspace{-1mm}
\end{table*}

\begin{table*}[t]
    \centering
    \tabcolsep=0.09cm
    \footnotesize
    \caption{Multi-class anomaly detection/localization on VisA with I-F1max/P-F1max metrics.}
    \label{tab:tab_visA_F1max}
    \begin{tabular}{c|c|c|c|c|c|c|c|c|c|c}
        \toprule
        \multicolumn{2}{c|}{Category} & UniAD \cite{uniad}& SimpleNet \cite{simplenet} & MambaAD \cite{mambaad}& DiAD \cite{diad} & GLAD \cite{glad} & ViTAD \cite{vitad}& AnomalyDINO & Costfilter-AD & Ours \\
        \midrule
       \multirow{4}{*}{\rotatebox{90}{\shortstack{Complex\\Structure}}} 
            & PCB1 & 87.8 / 8.3 & 86.0 / \textbf{78.8} & 91.6 / 72.4 & 80.7 / 52.8 & 70.1 / 44.4 & \textbf{91.8} / 61.7 & 82.2 / 68.0 & 91.6 / 66.9 & 85.7 / 74.1 \\
            & PCB2 & 83.1 / 9.2 & 84.5 / 18.6 & 89.3 / 23.4 & 84.7 / 16.7 & 83.3 / 14.4 & 85.3 / 21.2 & 76.2 / 23.7 & \textbf{92.0} / 21.1 & 85.7 / \textbf{35.3} \\
            & PCB3 & 76.1 / 21.9 & 82.6 / 36.1 & 86.7 / 27.4 & 77.6 / 18.8 & 87.6 / 27.7 & 83.9 / 26.4 & 80.2 / \textbf{38.8} & 82.1 / 28.6 & \textbf{90.0} / 34.2 \\
            & PCB4 & 94.3 / 22.9 & 93.5 / 32.9 & \textbf{98.5} / 46.9 & 97.0 / 27.2 & 98.0 / \textbf{52.0} & 96.6 / 48.3 & 91.0 / 30.8 & 97.1 / 39.4 & 92.0 / 45.4 \\
        \midrule
        \multirow{4}{*}{\rotatebox{90}{\shortstack{Multiple\\instances}}}
            & Macaroni 1 & 72.7 / 9.7 & 73.1 / 8.4 & 81.6 / \textbf{27.6} & 78.8 / 16.7 & 85.4 / 19.2 & 76.7 / 19.3 & 79.5 / 17.3 & 86.7 / 19.7 & \textbf{87.6} / 20.8 \\
            & Macaroni 2 & 69.9 / 4.3 & 59.7 / 3.9 & 73.8 / 16.1 & 69.6 / 2.8 & 71.8 / 19.3 & 74.9 / 10.4 & 73.0 / 11.1 & 81.5 / 15.5 & \textbf{83.5} / \textbf{22.9} \\
            & Capsules & 76.9 / 7.4 & 74.6 / 53.3 & 88.8 / \textbf{59.8} & 78.5 / 21.0 & 85.9 / 53.3 & 79.8 / 41.4 & 89.8 / 45.6 & 79.4 / 51.4 & \textbf{94.1} / 58.2 \\
            & Candle & 86.1 / 27.9 & 76.6 / 16.5 & \textbf{90.1} / 32.4 & 87.6 / 22.8 & 79.8 / 36.6 & 83.7 / 26.4 & 82.9 / 30.6 & 92.5 / 41.0 & 89.6 / \textbf{42.5 }\\
        \midrule
        \multirow{4}{*}{\rotatebox{90}{\shortstack{Single\\instances}}}
           & Cashew & 91.4 / 58.3 & 84.7 / 66.0 & 91.1 / 51.4 & 89.7 / 60.9 & 90.5 / 38.2 & 86.1 / 62.7 & \textbf{92.0} / 60.3 & 91.5 / 66.1 & \textbf{92.0} / \textbf{66.4} \\
            & \scriptsize{Chewing gum} & 95.2 / 56.1 & 93.8 / 29.8 & 94.2 / 59.9 & 95.9 / 25.8 & 95.5 / 69.6 & 91.4 / 58.7 & \textbf{97.5} / 56.9 & 96.9 / 64.7 & 97.0 / \textbf{72.3} \\
            & Fryum & 85.0 / 40.6 & 83.3 / 45.4 & 90.5 / 51.9 & 87.2 / \textbf{60.1} & 95.8 / 43.5 & 90.9 / 50.3 & \textbf{92.7} / 45.2 & 86.8 / 54.0 & 90.8 / 57.2 \\
            & Pipe fryum & 93.9 / 57.7 & 88.6 / 63.4 & 97.0 / 58.5 & 93.7 / 69.9 & 97.0 / 55.1 & 94.7 / 66.5 & 97.5 / 56.2 & 93.5 / \textbf{71.3} & \textbf{98.5} / 60.9 \\
        \midrule
        \multicolumn{2}{c|}{Mean} &84.4 / 27.0 &81.8 / 37.8 & 89.4 / 44.0 & 85.1 / 33.0& 86.7 / 39.4& 86.3 / 41.1& 86.2 / 40.4 & 89.3 / 45.0& \textbf{90.5} / \textbf{49.2} \\
        \bottomrule
    \end{tabular}
    \vspace{-1mm}
\end{table*}

\begin{table*}[t]
    \centering
    \tabcolsep=0.08cm
    \footnotesize
    \caption{Multi-class anomaly localization on VisA with AUPRO metrics.}
    \label{tab:tab_visA_PRO}
    \begin{tabular}{c|c|c|c|c|c|c|c|c|c|c}
        \toprule
        \multicolumn{2}{c|}{Category} & UniAD \cite{uniad}& SimpleNet \cite{simplenet} & MambaAD \cite{mambaad}& DiAD \cite{diad} & GLAD \cite{glad} & ViTAD \cite{vitad}& AnomalyDINO \cite{Anomalydino} & Costfilter-AD \cite{CostFilterAD} & Ours \\
        \midrule
       \multirow{4}{*}{\rotatebox{90}{\shortstack{Complex\\Structure}}} 
            & PCB1 & 64.1 & 83.6 & \textbf{92.8} & 80.2 & 88.3 & 89.6 & 82.8 & 88.3 & 90.2 \\
            & PCB2 & 66.9 & 85.7 & \textbf{89.6} & 67.0 & 91.7 & 82.0 & 77.7 & 85.4 & 78.5 \\
            & PCB3 & 70.6 & 85.1 & 89.1 & 68.9 & \textbf{94.2} & 88.0 & 79.7 & 75.6 & 82.0 \\
            & PCB4 & 72.3 & 61.1 & 87.6 & 85.0 & \textbf{94.9} & 91.8 & 83.1 & 84.4 & 86.5 \\
        \midrule
        \multirow{4}{*}{\rotatebox{90}{\shortstack{Multiple\\Instances}}}
            & Macaroni 1 & 84.0 & 92.0 & 95.2 & 68.5 & \textbf{99.1} & 89.2 & 90.2 & 96.4 & 98.0 \\
            & Macaroni 2 & 76.6 & 77.8 & 96.2 & 73.1 & 97.2 & 87.2 & 84.8 & 94.2 & \textbf{97.5} \\
            & Capsules & 43.7 & 73.7 & 91.8 & 77.9 & 91.8 & 75.1 & 86.1 & 61.9 & \textbf{90.1} \\
            & Candle & 91.6 & 87.6 & 95.5 & 89.4 & 92.8 & 85.2 & 94.1 & 95.2 & \textbf{96.9} \\
        \midrule
        \multirow{4}{*}{\rotatebox{90}{\shortstack{Single\\Instances}}}
           & Cashew & 87.9 & 84.1 & 87.8 & 61.8 & 61.1 & 78.8 & 91.3 & 90.5 & \textbf{96.3} \\
            & \scriptsize{Chewing gum} & 81.3 & 78.3 & 79.7 & 59.5 & 92.5 & 71.5 & 85.7 & 88.5 & \textbf{94.5} \\
            & Fryum & 76.2 & 85.1 & 91.6 & 81.3 & \textbf{96.4} & 87.8 & 85.0 & 86.9 & 92.3 \\
            & Pipe fryum & 91.5 & 83.0 & 95.1 & 89.9 & \textbf{98.0} & 94.7 & 94.7 & 94.5 & 97.1 \\
        \midrule
        \multicolumn{2}{c|}{Mean} &75.6 &81.4 &91.0 & 75.2& 91.5 & 85.1 & 86.3 & 86.8 & \textbf{91.7} \\
        \bottomrule
    \end{tabular}
    \vspace{-1mm}
\end{table*}

\section{Extended Visualization Results}\label{supp:results3}
\begin{figure*}
\begin{center}
\centerline{\includegraphics[width=0.99\textwidth]{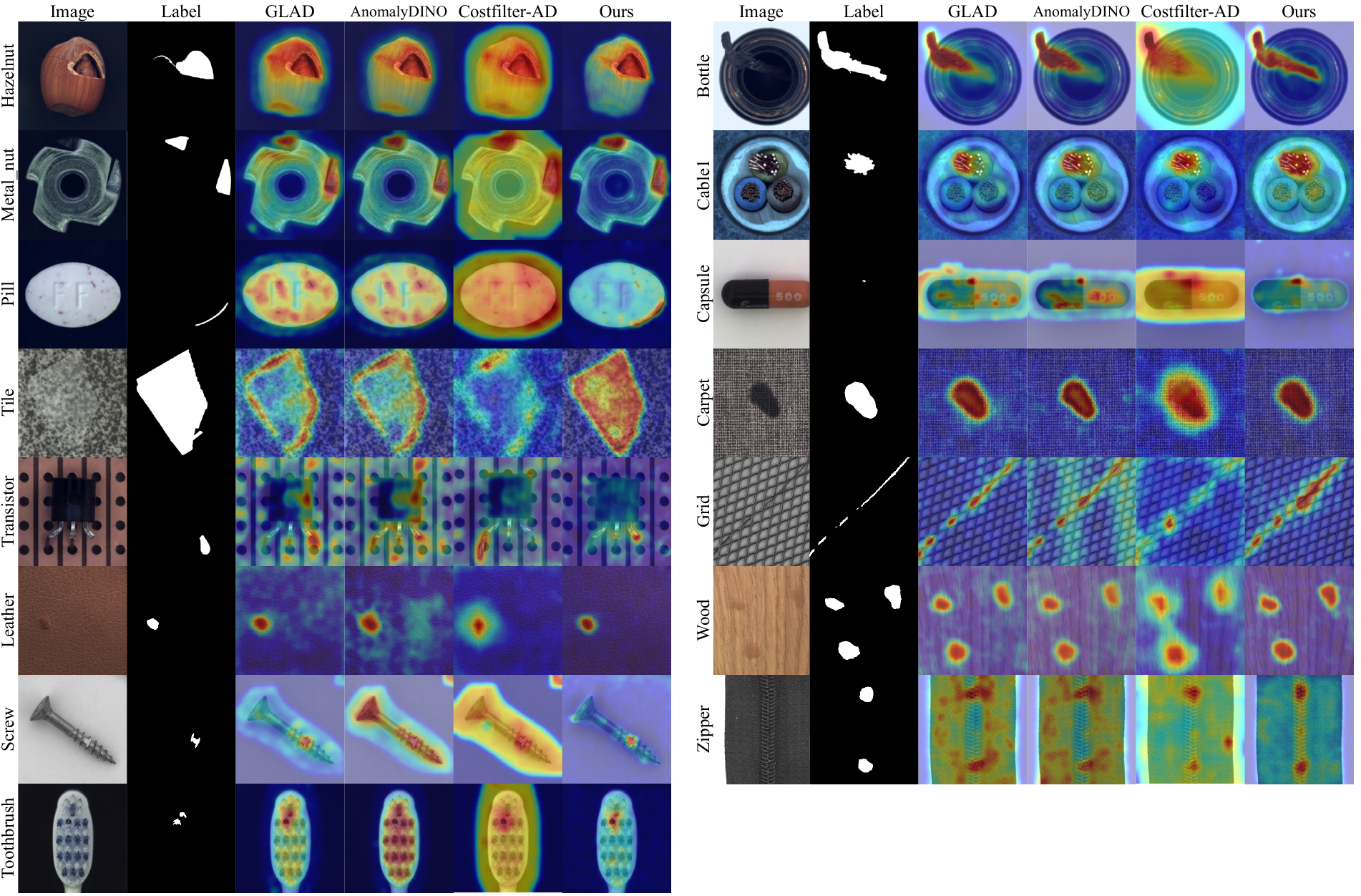}}
\vspace{-2mm}
\caption{Qualitative comparison of multi-class anomaly localization results on MVTec-AD dataset.}
\label{fig:qualitative_mvtec}
\end{center}
\vspace{-8mm}
\end{figure*}

\begin{figure*}
\begin{center}
\centerline{\includegraphics[width=0.99\textwidth]{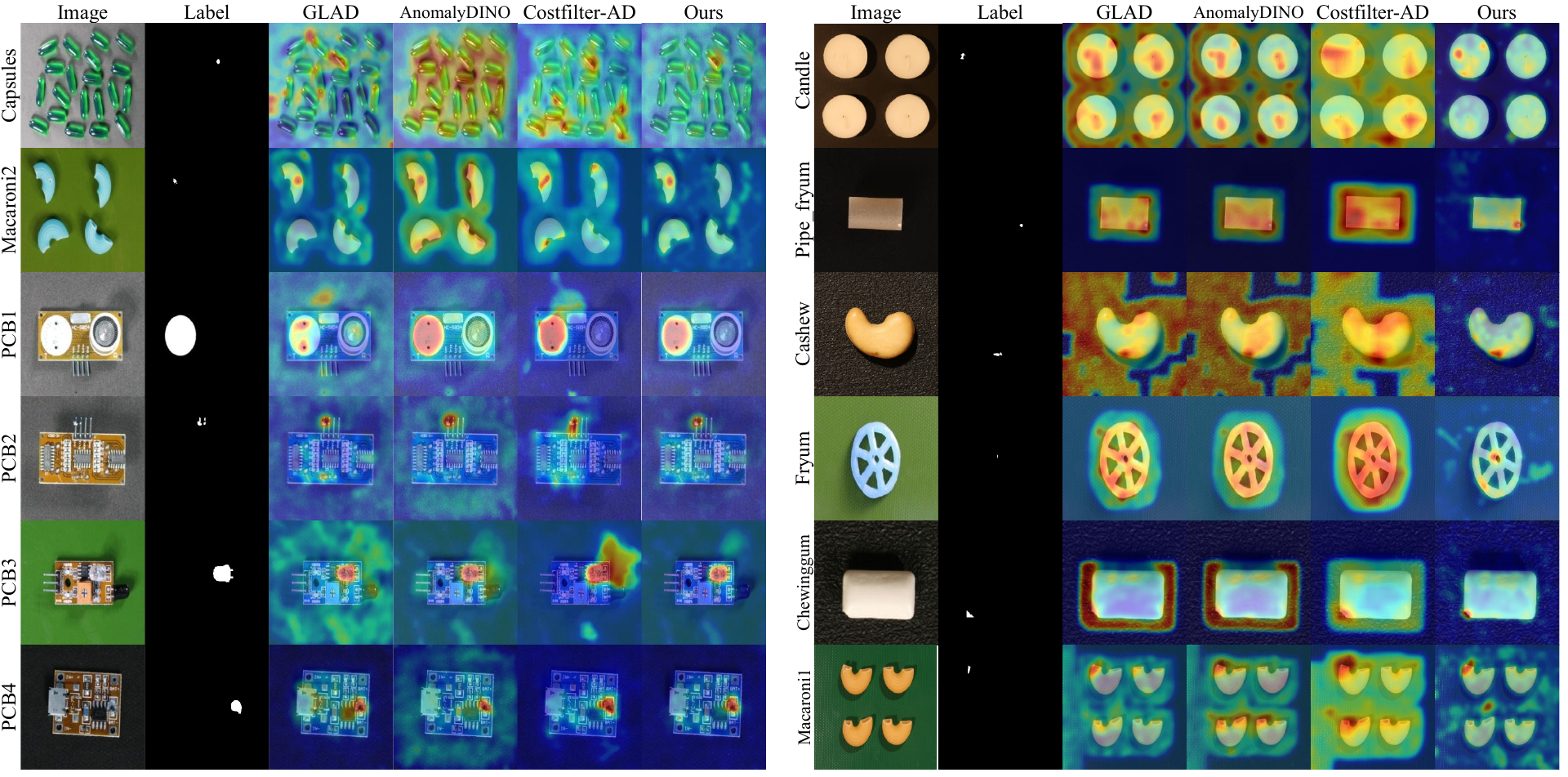}}
\vspace{-2mm}
\caption{Qualitative comparison of multi-class anomaly localization results on VisA dataset.}
\label{fig:qualitative_visa}
\end{center}
\vspace{-8mm}
\end{figure*}

To further demonstrate the effectiveness of our approach, we provide multi-class UAD qualitative visualizations of anomaly detection heatmaps on MVTec-AD and VisA in Fig.~\ref{fig:qualitative_mvtec} and Fig.~\ref{fig:qualitative_visa}. Compared with representative baselines, our method produces cleaner and more structurally coherent anomaly maps, exhibiting reduced noise, fewer spurious activations, and sharper defect boundaries. These results highlight the ability of our RAID model to deliver reliable and fine-grained localization, even for subtle or low-contrast anomalies.

\begin{table*}[t]
    \centering
    \caption{\textbf{Few-shot} anomaly detection and localization performance of RAID on MVTec-AD, with per-category I-AUROC and P-AUROC reported as mean (\%) $\pm$ standard deviation (\%).}
    \label{tab:few-shot-mvtec}
    \vspace{-2mm}
    \footnotesize
\setlength{\tabcolsep}{17pt}
    \begin{tabular}{ll c c c c c c }
        \toprule
         \multirow{2}{*}{Object} &  \multicolumn{2}{c}{1-shot} & \multicolumn{2}{c}{2-shot} &  \multicolumn{2}{c}{4-shot}\\
\cmidrule(lr){2-3} \cmidrule(lr){4-5} \cmidrule(lr){6-7}
&I-AUROC & P-AUROC & I-AUROC & P-AUROC & I-AUROC & P-AUROC \\
\midrule
 Bottle  &99.9$\scriptstyle \pm 0.1$ & 99.2$\scriptstyle \pm 0.1$ & 100.0$\scriptstyle \pm 0.0$ & 99.3$\scriptstyle \pm 0.1$ & 100.0$\scriptstyle \pm 0.1$ & 99.3$\scriptstyle \pm 0.0$

\\
 Cable & 93.5$\scriptstyle \pm 1.2$ & 92.7$\scriptstyle \pm 1.4$ & 95.3$\scriptstyle \pm 0.8$ & 94.0$\scriptstyle \pm 0.5$ & 95.8$\scriptstyle \pm 1.1$ &94.5$\scriptstyle \pm 0.4$
\\
 Capsule  &79.7$\scriptstyle \pm 3.9$ & 98.5$\scriptstyle \pm 0.2$ &85.8$\scriptstyle \pm 7.6$ & 98.7$\scriptstyle \pm 0.2$ & 91.0$\scriptstyle \pm 2.7$ & 98.9$\scriptstyle \pm 0.1$
\\
 Hazelnut & 93.5$\scriptstyle \pm 12.1$ & 99.0$\scriptstyle \pm 1.0$ & 98.0$\scriptstyle \pm 2.9$ & 99.5$\scriptstyle \pm 0.2$ & 99.9$\scriptstyle \pm 0.2$ & 99.6$\scriptstyle \pm 0.1$\\
 Metal\_nut &99.8$\scriptstyle \pm 0.2$ & 96.3$\scriptstyle \pm 0.4$ & 99.9 $\scriptstyle \pm0.1$ & 96.4$\scriptstyle \pm 0.5$ & 100.0$\scriptstyle \pm 0.0$ & 97.0$\scriptstyle \pm 0.3$
\\
Pill & 94.5$ \scriptstyle \pm 1.2$ & 97.5$ \scriptstyle \pm 0.3$ & 96.2$ \scriptstyle \pm 0.9$ & 98.0$ \scriptstyle \pm 0.2$ & 97.2$ \scriptstyle \pm 0.3$ & 98.0$ \scriptstyle \pm 0.3$
\\
 Screw  & 84.3$ \scriptstyle \pm 3.0$& 96.8$ \scriptstyle \pm 0.7$ & 83.3$ \scriptstyle \pm 2.3$ & 95.3$ \scriptstyle \pm  0.3$ & 80.2$ \scriptstyle \pm  3.0$ & 93.2$ \scriptstyle \pm  0.9$
\\
 Toothbrush  &97.8$ \scriptstyle \pm  0.5$  & 99.1$ \scriptstyle \pm  0.3$ & 98.6$ \scriptstyle \pm  0.9$ & 99.3$ \scriptstyle \pm  0.1$ & 99.6$ \scriptstyle \pm  0.5$ & 99.4$ \scriptstyle \pm  0.1$
 \\
 Transisitor &87.3$ \scriptstyle \pm  7.1$ & 86.4$ \scriptstyle \pm  6.5$ & 94.2$ \scriptstyle \pm  2.1$ & 91.7$ \scriptstyle \pm  0.5$ & 92.7$ \scriptstyle \pm  5.0$ & 90.0$ \scriptstyle \pm  3.7$

\\
 Zipper & 97.9$ \scriptstyle \pm  0.5$& 92.2$ \scriptstyle \pm  2.0$ & 98.3$ \scriptstyle \pm  0.7$ & 92.7$ \scriptstyle \pm  1.0$ & 98.1$ \scriptstyle \pm  0.3$ & 93.2$ \scriptstyle \pm  0.8$

 \\
 Carpet  &100.0$ \scriptstyle \pm  0.0$& 99.4$\scriptstyle \pm 0.1$ & 100.0$\scriptstyle \pm 0.0$ & 99.4$\scriptstyle \pm 0.1$ & 100.0$\scriptstyle \pm 0.0$ & 99.4$\scriptstyle \pm 0.1$

\\
 Grid  & 99.2$\scriptstyle \pm 1.1$  & 99.1$\scriptstyle \pm 0.1$ & 99.9$\scriptstyle \pm 0.1$ & 99.2$\scriptstyle \pm 0.1$ & 99.9$\scriptstyle \pm 0.1$ & 99.2$\scriptstyle \pm 0.1$
\\
 Leather  & 100.0$\scriptstyle \pm 0.0$ & 99.0$\scriptstyle \pm 0.1$ & 100.0$\scriptstyle \pm 0.0$ & 98.9$\scriptstyle \pm 0.1$ & 100.0$\scriptstyle \pm 0.1$ & 98.8$\scriptstyle \pm 0.1$

\\
 Tile  & 100.0$\scriptstyle \pm 0.0$ & 97.0$\scriptstyle \pm 0.2$ & 100.0$\scriptstyle \pm 0.0$ & 97.1$\scriptstyle \pm 0.2$ & 100.0$\scriptstyle \pm 0.0$ & 97.1$\scriptstyle \pm 0.1$

\\
 Wood  & 97.9$\scriptstyle \pm 0.7$ & 96.5$\scriptstyle \pm 0.5$ & 98.8$\scriptstyle \pm 0.5$ & 96.7$\scriptstyle \pm 0.4$ & 98.9$\scriptstyle \pm 0.3$ & 96.4$\scriptstyle \pm 0.7$
\\
\midrule
 Mean  &95.1$\scriptstyle \pm 0.7$ & 96.6$\scriptstyle \pm 0.5$ & 96.6$\scriptstyle \pm 0.7$ & 97.1$\scriptstyle \pm 0.1$ & 96.9$\scriptstyle \pm 0.3$ & 96.9$\scriptstyle \pm 0.3$
\\

    \midrule
    \end{tabular}
    \vspace{-4mm}
\end{table*}

\begin{table*}[t]
    \centering
    \caption{\textbf{Few-shot} anomaly detection and localization performance of RAID on VisA, with per-category I-AUROC and P-AUROC reported as mean (\%) $\pm$ standard deviation (\%).}
    \label{tab:few-shot_visa}
    \vspace{-2mm}
    \footnotesize
\setlength{\tabcolsep}{17pt}
    \begin{tabular}{ll c c c c c c }
        \toprule
         \multirow{2}{*}{Object} &  \multicolumn{2}{c}{1-shot} & \multicolumn{2}{c}{2-shot} &  \multicolumn{2}{c}{4-shot}\\
\cmidrule(lr){2-3} \cmidrule(lr){4-5} \cmidrule(lr){6-7}
&I-AUROC & P-AUROC & I-AUROC & P-AUROC & I-AUROC & P-AUROC \\
\midrule
 PCB1  &72.1$\scriptstyle \pm 8.9$ & 97.4$\scriptstyle \pm 0.2$ & 79.5$\scriptstyle \pm 0.9$ & 97.9$\scriptstyle \pm 0.3$& 78.1$\scriptstyle \pm 2.6$ & 98.2$\scriptstyle \pm 0.1$
\\
 PCB2  & 81.4$\scriptstyle \pm 2.0$ & 97.2$\scriptstyle \pm 0.5$ & 81.5$\scriptstyle \pm 4.1$ & 97.5$\scriptstyle \pm 0.2$ &83.8$\scriptstyle \pm 1.4$ & 97.4$\scriptstyle \pm 0.3$
\\
 PCB3  &86.6$\scriptstyle \pm 1.1$ & 96.6$\scriptstyle \pm 0.1$ & 89.0$\scriptstyle \pm 3.0$ & 97.2$\scriptstyle \pm 0.2$ &89.0$\scriptstyle \pm 1.6$ & 98.0$\scriptstyle \pm 0.1$

\\
 PCB4 & 93.4$\scriptstyle \pm 1.2$ & 96.7$\scriptstyle \pm 0.7$ & 96.4$\scriptstyle \pm 0.9$ & 97.3$\scriptstyle \pm 0.2$ & 95.8$\scriptstyle \pm 1.2$ & 97.4$\scriptstyle \pm 0.3$
\\

Macaroni~1 & 85.1$\scriptstyle \pm 5.0$ & 97.2$\scriptstyle \pm 1.2$ & 84.3$\scriptstyle \pm 3.0$ & 96.2$\scriptstyle \pm 1.9$ &87.7$\scriptstyle \pm 2.1$ & 96.4$\scriptstyle \pm 1.8$

\\
 Macaroni~2 & 59.5$\scriptstyle \pm 4.9$ & 97.5$\scriptstyle \pm 0.5$ & 70.0$\scriptstyle \pm 6.3$ & 98.0$\scriptstyle \pm 0.4$ &70.6$\scriptstyle \pm 6.7$ & 97.9$\scriptstyle \pm 0.5$

\\
 Capsules  &95.0$\scriptstyle \pm 2.6$ & 98.7$\scriptstyle \pm 0.1$ &98.0$\scriptstyle \pm 1.0$ & 99.0$\scriptstyle \pm 0.1$ & 95.6$\scriptstyle \pm 1.7$ & 98.6$\scriptstyle \pm 0.3$

\\
 Candle  & 87.6$\scriptstyle \pm 1.8$ & 98.5$\scriptstyle \pm 0.1$ & 88.8$\scriptstyle \pm 0.4$ &98.9$\scriptstyle \pm 0.1$ & 92.2$\scriptstyle \pm 0.5$ & 99.2$\scriptstyle \pm 0.0$

 \\
 
 Cashew   & 92.0$\scriptstyle \pm 3.3$ & 98.3$\scriptstyle \pm 0.4$ & 92.6$\scriptstyle \pm 2.0$ & 98.0$\scriptstyle \pm 0.2$ &93.5$\scriptstyle \pm 0.9$ & 99.2$\scriptstyle \pm 0.2$

\\
 Chewing gum  & 97.1$\scriptstyle \pm 0.6$ & 98.9$\scriptstyle \pm 0.1$ & 97.6$\scriptstyle \pm 0.9$ &98.8$\scriptstyle \pm 0.1$ & 97.2$\scriptstyle \pm 0.2$ & 98.9$\scriptstyle \pm 0.1$

\\ 
 Fryum  & 91.9$\scriptstyle \pm 1.9$ & 96.6$\scriptstyle \pm 0.5$ & 92.9$\scriptstyle \pm 1.8$ & 96.9$\scriptstyle \pm 0.4$ & 93.6$\scriptstyle \pm 1.4$ & 97.5$\scriptstyle \pm 0.1$

\\
 Pipe fryum  & 87.4$\scriptstyle \pm 5.8$ & 98.8$\scriptstyle \pm 0.3$ & 90.9$\scriptstyle \pm 4.7$ & 99.2$\scriptstyle \pm 0.0$ & 95.5$\scriptstyle \pm 0.7$ & 99.4$\scriptstyle \pm 0.1$

\\
\midrule
 Mean  & 85.8$\scriptstyle \pm 0.6$ & 97.1$\scriptstyle \pm 0.1$ & 88.5$\scriptstyle \pm 0.2$ &97.9$\scriptstyle \pm 0.2$ & 89.3$\scriptstyle \pm 0.5$ & 98.2$\scriptstyle \pm 0.2$
\\
    \midrule
 
    \end{tabular}
    \vspace{-4mm}
\end{table*}

\begin{figure*}
\begin{center}
\centerline{\includegraphics[width=0.999\textwidth]{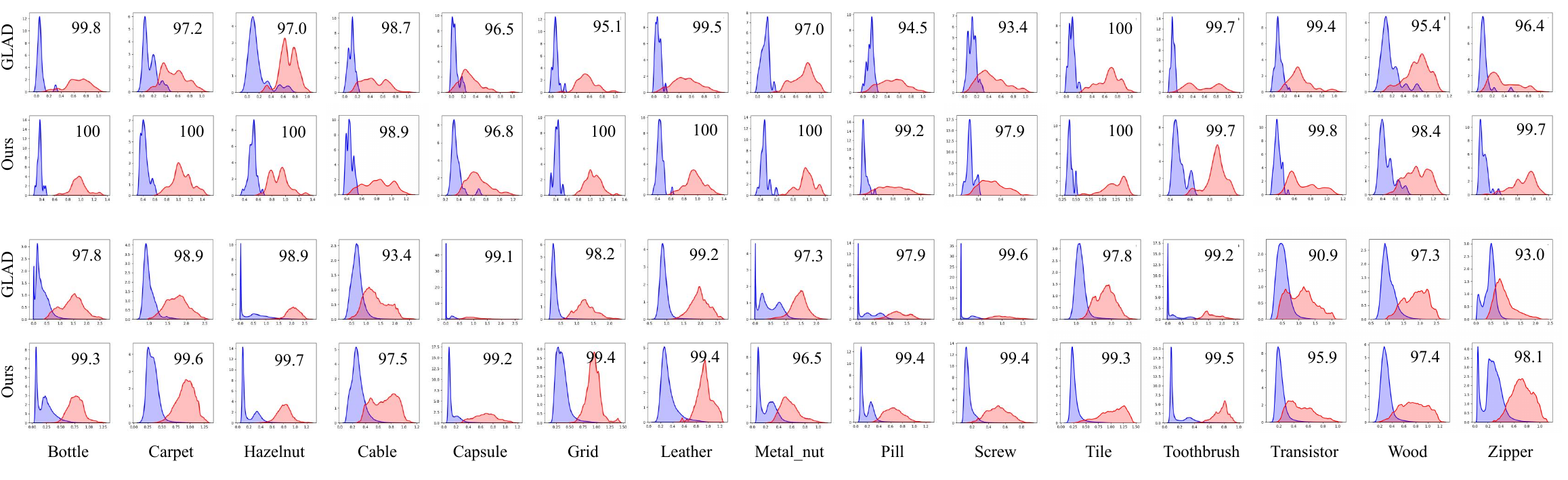}}
\vspace{-2mm}
\caption{KDE curves of image-level and pixel-level anomaly scores on MVTec-AD. The first two rows correspond to image-level metrics, while the last two rows correspond to pixel-level metrics.}
\label{fig:KDE_mvtec}
\end{center}
\vspace{-8mm}
\end{figure*}

\begin{figure*}
\begin{center}
\centerline{\includegraphics[width=0.85\textwidth]{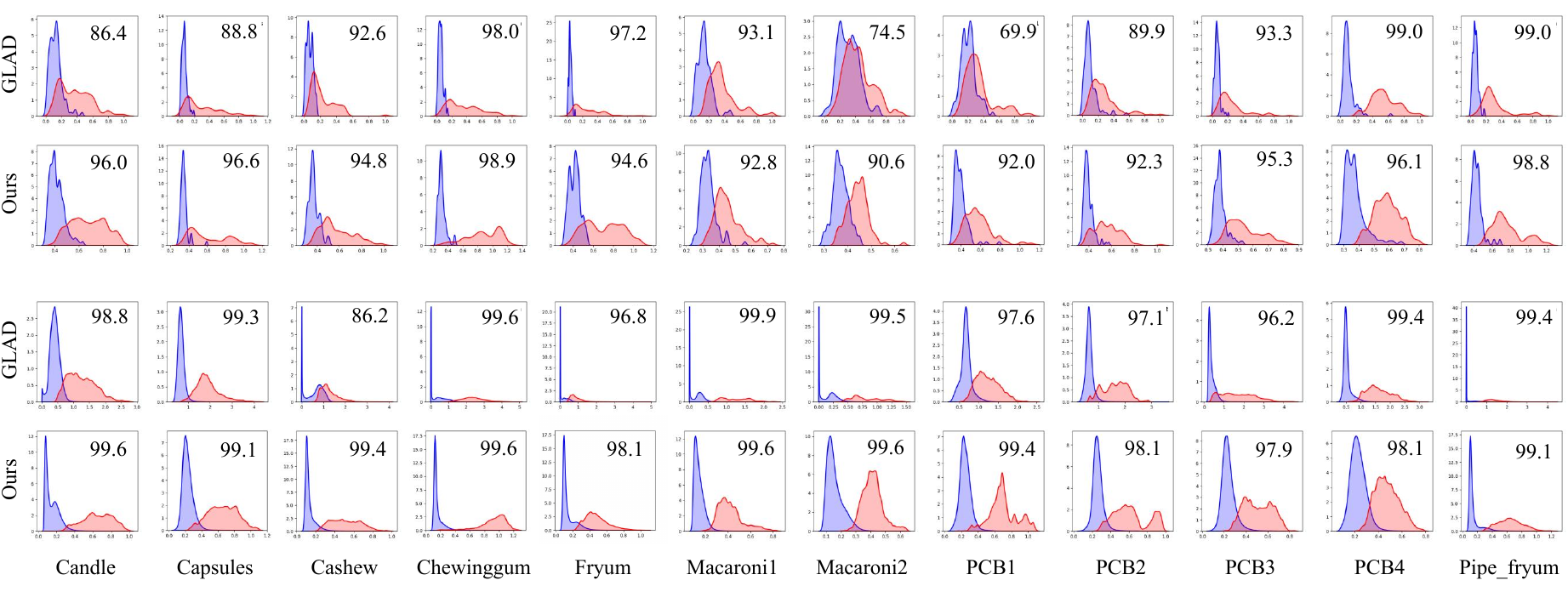}}
\vspace{-2mm}
\caption{KDE curves of image-level and pixel-level anomaly scores on VisA. The first two rows correspond to image-level metrics, while the last two rows correspond to pixel-level metrics.}
\label{fig:KDE_visa}
\end{center}
\vspace{-8mm}
\end{figure*}

To further demonstrate the effectiveness of RAID, we analyze the distributions of anomaly scores using KDE. Fig.~\ref{fig:KDE_mvtec} and Fig.~\ref{fig:KDE_visa} show the KDE curves for MVTec-AD and VisA, respectively, where the first two rows correspond to image-level metrics and the last two rows correspond to pixel-level metrics. Pink and blue represent abnormal and normal samples. Compared with a representative reconstruction-based method (GLAD~\cite{glad}), RAID yields substantially reduced overlap between normal and abnormal score distributions, indicating stronger discriminative power at both the image and pixel levels.

\begin{figure*}
\begin{center}
\centerline{\includegraphics[width=0.9\textwidth]{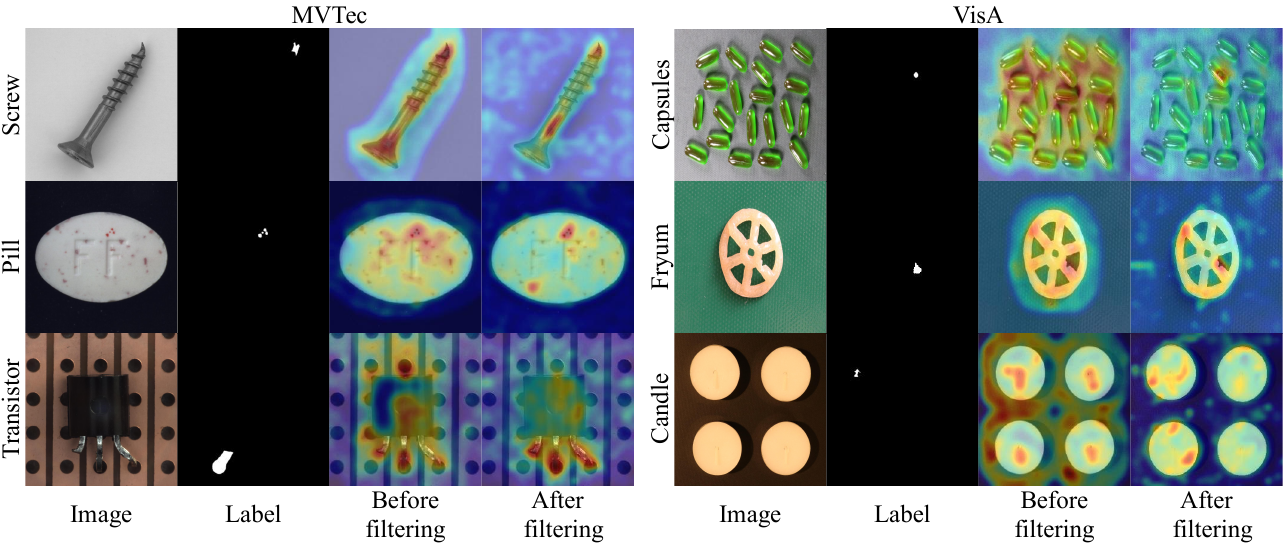}}
\vspace{-2mm}
\caption{Failure cases on MVTec-AD and VisA. Extremely subtle or low-contrast anomalies may still cause missed or imprecise localization, likely due to insufficient discriminative cues encoded in the cost volume.}
\label{fig:failure_case}
\end{center}
\vspace{-2mm}
\end{figure*}

\section{Failure Cases Analysis}\label{supp:fail}
As shown in Fig.~\ref{fig:failure_case}, although the guided MoE filter effectively suppresses matching noise, failures still arise when the cost volume provides weak or unreliable anomaly cues. Typical examples include reflections in the \textit{Screw} case, irregular speckled patterns in the \textit{Pill} case, and textureless surfaces in the challenging \textit{Candle} case. While RAID can detect subtle anomalies such as the small spot in the \textit{Capsules} case, low-resolution inputs and smooth textures often weaken feature discriminability, leading to ambiguous retrieval. This ambiguity may cause the filter to oversuppress true anomalies or retain residual noise. Addressing such visually ambiguous scenarios remains an important challenge for future work.

\end{document}